\documentclass{article}
\usepackage{wrapfig}
\usepackage{graphicx} 
\usepackage{url}
\title{AI Agents: Evolution, Architecture, and Real-World Applications}
\author{Naveen Krishnan}
\date{March 2025}

\begin{document}

\maketitle

\section{Introduction}

\bigskip

Artificial Intelligence (AI) has evolved dramatically over the past decade, transitioning from specialized systems
designed for narrow tasks to increasingly sophisticated architectures capable of autonomous operation across diverse
domains. Among these advancements, AI agents represent a particularly significant development, embodying a paradigm
shift in how intelligent systems interact with their environments, make decisions, and achieve complex goals. Unlike
traditional AI systems that execute predefined algorithms within constraints, AI agents possess the capacity
to autonomously perceive, reason, and act, often adapting their behavior based on environmental feedback and
accumulated experience.

\bigskip

The concept of an AI agent refers to a system or program that is capable of autonomously performing tasks on behalf of a
user or another system by designing its workflow and utilizing available tools. These agents can encompass a wide range
of functionalities beyond natural language processing, including decision making, problem solving, interacting with external environments, and executing actions. As Kapoor et al. (2024) note in their analysis of agent benchmarks, the development of AI agents represents an exciting new research direction with significant implications for real-world applications across numerous industries.

\bigskip

The evolution of AI agents has been accelerated by recent breakthroughs in large language models (LLMs), which have
provided a foundation for more sophisticated reasoning capabilities. Modern AI agents leverage these advanced language
models as core components, augmenting them with specialized modules for memory, planning, tool use, and environmental
interaction. This integration enables agents to perform complex tasks that would be challenging or impossible for
traditional AI systems, from reconciling financial statements to providing step-by-step instructions for field
technicians based on contextual understanding of product information.

\bigskip

The distinction between AI agents and other AI systems lies primarily in their architecture and operational
capabilities. While conventional AI systems typically operate within predefined parameters and require explicit instructions for each task, AI agents demonstrate greater autonomy in goal-directed behavior. As described by Microsoft's chief marketing officer for AI at Work, Jared Spataro, Agents are like layers on top of the
language models that observe and collect information, provide input to the model and together generate an action plan (Ray, 2024). This architectural approach allows agents to decompose complex problems into manageable subtasks, reason over available information, utilize appropriate tools, and learn from feedback while maintaining context across interactions.

\bigskip

The taxonomy of AI agents has evolved to include various categories based on their cognitive capabilities and
operational mechanisms. AWS (2024) identifies several distinct types, including simple reflex agents that operate based
on predefined rules, model-based agents that evaluate probable outcomes before deciding, goal-based agents that compare
different approaches to achieve desired outcomes, utility-based agents that maximize specific value metrics, learning
agents that continuously adapt based on experience, and hierarchical agents that coordinate across multiple levels of
abstraction. Each type represents a different approach to agent design, with corresponding strengths and limitations
for particular use cases.

\bigskip

The current landscape of research and development of AI Agents is characterized by rapid innovation across academic
institutions and industry leaders. Organizations like IBM, Microsoft, AWS, and Anthropic are developing increasingly
sophisticated agent architectures, while research teams at institutions such as Princeton University are establishing
new benchmarks and evaluation frameworks. This convergence of academic and commercial interest has accelerated progress
in the field, although challenges remain in areas such as reasoning capabilities, tool integration, and evaluation
methodologies.

\bigskip

Despite significant advances, the development of AI agents faces several critical challenges. As highlighted by Kapoor
et al. (2024), current agent benchmarks often focus narrowly on accuracy without sufficient attention to other
important metrics such as cost-effectiveness, reproducibility, and real-world applicability. Additionally, the
conflation of benchmarking needs between model developers and downstream application developers has complicated efforts
to identify which agent architectures are best suited for particular use cases. These challenges underscore the need
for more sophisticated evaluation frameworks that can capture the multidimensional nature of agent performance in diverse contexts.

\bigskip

This paper aims to provide a comprehensive analysis of AI agents, examining their theoretical foundations, architectural
components, evaluation methodologies, and real-world applications. We begin by reviewing the literature on agent systems, tracing the evolution of key concepts and frameworks that have shaped current approaches. We then explore the core components of modern agent architectures, including perception mechanisms, knowledge representation, reasoning modules, and action selection. Next, we examine current evaluation practices and propose a more holistic framework for assessing agent performance. We then analyze real-world applications across enterprise contexts, personal assistance, and specialized domains, highlighting both successes and limitations. Finally, we discuss ongoing challenges in the field and identify promising directions for future research.

\bigskip

By synthesizing insights from both academic research and industry implementations, this paper contributes to the growing
literature on AI agents in several ways. First, it provides a unified conceptual framework for understanding diverse
agent architectures and their capabilities. Second, it critically examines current evaluation methodologies and
proposes improvements that better align with real-world requirements. Third, it offers a detailed analysis of
successful agent applications across multiple domains, extracting generalizable principles for effective deployment.
Finally, it identifies key research challenges and opportunities that will shape the future development of agent
systems. Through these contributions, our aim is to advance both theoretical understanding and practical implementation of
AI agents as they continue to transform how intelligent systems operate in the world.

\bigskip

\section{ Literature Review}

\bigskip

\subsection{Theoretical Foundations of AI Agents}

\bigskip

The concept of artificial intelligence agents has deep roots in computer science, philosophy, and cognitive science. The
theoretical underpinnings of agent-based systems can be traced back to early work on distributed artificial intelligence in the 1970s and 1980s. However, the modern conceptualization of AI agents emerged more distinctly in the 1990s, with seminal contributions from researchers like Russell and Norvig (1995), who defined an agent as
anything that can be viewed as perceiving its environment through sensors and acting upon that environment through actuators. This definition established the fundamental perception-action loop that
characterizes agent systems.

\bigskip

Wooldridge and Jennings (1995) further refined the concept by proposing that intelligent agents should possess several
key properties: autonomy (operating without direct human intervention), social ability (interacting with other agents
and humans), reactivity (responding to environmental changes), and proactivity (exhibiting goal-directed behavior).
These properties have remained central to agent theory, though their implementation has evolved significantly with
advances in computational capabilities and algorithmic approaches.

\bigskip

Agency theory in artificial intelligence draws from multiple disciplines, including economics, cognitive psychology, and
philosophy of mind. The concept of agency implies a relationship in which one entity (the agent) acts on behalf of
another (the principal), making decisions that affect outcomes relevant to the principal's interests. In AI contexts,
this relationship is formalized through goal specifications, utility functions, and constraints that guide agent behavior. The philosophical dimensions of agency including questions about intention, belief, and rationality have informed computational models of agent reasoning and decision-making.

\bigskip

The notion of rationality has been particularly influential in agent design. As articulated by Russell and Norvig
(2010), a rational agent is one that acts so as to achieve the best outcome or, when there is
uncertainty, the best expected outcome. This principle has guided the development of decision-theoretic
approaches to agent architecture, where actions are selected based on their expected utility given the agent's current
knowledge state. However, as noted by Kahneman and Tversky (1979) in their work on prospect theory, human
decision-making often deviates from strict rationality in predictable waysinsights that have
informed more nuanced models of agent behavior that incorporate bounded rationality and satisficing.

\bigskip

The concept of autonomy in agent systems has also evolved significantly. Early definitions emphasized independence from
direct human control, but contemporary perspectives recognize degrees of autonomy along multiple dimensions. As
Bradshaw et al. (2013) argue, autonomy should be understood as a relationship of interdependence rather than absolute independence. This re-conceptualization has important implications
for human-agent interaction, suggesting that effective agent design should focus on complementary capabilities and
appropriate trust calibration rather than maximizing agent independence.

\bigskip

Intelligence in agent systems has been conceptualized through various theoretical lenses. The symbolic approach,
influenced by the physical symbol system hypothesis of Newell and Simon (1976), models intelligence as manipulation of
symbolic representations according to formal rules. In contrast, the connectionist approach, exemplified by neural
network architectures, emphasizes emergent intelligence through distributed processing across networks of simple units.
Hybrid approaches that integrate symbolic reasoning with statistical learning have become increasingly prevalent,
particularly as large language models have demonstrated capabilities for both pattern recognition and symbolic
manipulation.

\bigskip

Recent theoretical work has expanded the conceptual foundations of agent systems to address challenges of scale,
complexity, and social interaction. Multi-agent system theories examine how collections of agents can coordinate their
actions to achieve goals beyond the capabilities of individual agents. Theories of emergent intelligence explore how
complex behaviors can arise from interactions among relatively simple agents. Social cognition theories inform the
design of agents that can model the mental states of other agents (including humans) and adapt their behavior
accordingly.

\bigskip

The evolution of theoretical frameworks for AI agents has been significantly influenced by advances in related fields.
Reinforcement learning theory has provided formal models for how agents can learn optimal policies through
environmental interaction and reward signals. Cognitive architectures like ACT-R (Anderson et al., 2004) and SOAR
(Laird, 2012) have offered integrated theories of how different cognitive processesperception,
memory, reasoning, learningcan be combined in unified agent architectures. More recently,
theories of predictive processing and active inference (Friston, 2010) have suggested that agent cognition can be
understood as continuous prediction of sensory inputs and selection of actions that minimize prediction error.

\bigskip

As AI agent technology has matured, theoretical perspectives have increasingly addressed ethical and social dimensions.
Value alignment theory examines how agent objectives can be designed to align with human values and preferences.
Theories of explainable AI consider how agent reasoning can be made transparent and interpretable to human users.
Theories of human-agent teamwork explore how agents and humans can effectively collaborate, with complementary
capabilities and shared mental models.

\bigskip

The theoretical landscape of AI agents continues to evolve, with new perspectives emerging from interdisciplinary
research at the intersection of computer science, cognitive science, neuroscience, philosophy, and social psychology.
These diverse theoretical foundations provide conceptual frameworks for understanding agent capabilities, limitations,
and potential, informing both technical implementation and responsible deployment.

\bigskip

\subsection{Typology of AI Agents}

\bigskip

The classification of AI agents has evolved significantly as the field has matured, with various taxonomies proposed to
categorize different agent architectures based on their capabilities, design principles, and operational
characteristics. Understanding these classifications provides a framework for analyzing the strengths, limitations, and
appropriate applications of different agent types.

\bigskip

Simple reflex agents represent the most basic agent architecture, operating strictly based on predefined rules and
immediate perceptual data. These agents implement condition-action rules (if-then statements) that map specific
environmental states to corresponding actions. As described by AWS (2024), simple reflex agents will not
respond to situations beyond a given event condition action rule, making them suitable for
straightforward tasks in stable, fully observable environments. Examples include thermostat controllers, basic chatbots
that detect specific keywords to trigger responses, and simple automated systems for password resets. While limited in
their capabilities, simple reflex agents offer advantages in terms of predictability, computational efficiency, and
ease of implementation.

\bigskip

Model-based reflex agents extend the capabilities of simple reflex agents by incorporating an internal model of the
world. This model allows the agent to maintain state information that is not directly observable in the current
environment, enabling more sophisticated decision-making. As noted by Russell and Norvig (2010), model-based agents can
keep track of the current state of the world and use this information to evaluate
probable outcomes before selecting actions. This approach is particularly valuable in partially observable environments
where the current percept alone is insufficient for optimal decision-making. Examples include navigation systems that
maintain maps of their environment, recommendation systems that build user preference models, and diagnostic systems
that infer internal states from observable symptoms.

\bigskip

Goal-based agents, also known as rule-based agents, incorporate explicit representations of desirable world states
(goals) and select actions specifically to achieve these goals. Unlike reflex agents that simply react to environmental
stimuli, goal-based agents engage in means-end reasoning, comparing different approaches to determine which will best
achieve their objectives. AWS (2024) notes that these agents always choose the most efficient
path and are suitable for performing complex tasks, such as natural language processing
(NLP) and robotics applications. The incorporation of goals enables these agents to exhibit more
flexible behavior across diverse situations, as the same goal can be achieved through different action sequences
depending on environmental conditions.

\bigskip

Utility-based agents refine the goal-based approach by introducing a utility function that assigns values to different
world states, allowing the agent to make more nuanced decisions when multiple goals conflict or when there are varying
degrees of goal satisfaction. Rather than simply distinguishing between goal states and non-goal states, utility-based
agents can evaluate the relative desirability of different outcomes. As AWS (2024) explains, these agents
compare different scenarios and their respective utility values or benefits and choose
actions that provide users with the most rewards. This approach is particularly valuable
for decision-making under uncertainty, where agents must balance multiple objectives or optimize across competing
criteria. Examples include financial trading systems, resource allocation algorithms, and travel planning assistants
that optimize across multiple preferences (e.g., cost, time, comfort).

\bigskip

Learning agents represent a significant advancement in agent architecture, incorporating mechanisms to improve
performance through experience. These agents modify their behavior based on feedback, gradually refining their internal
models, decision rules, or utility functions to better achieve their objectives. AWS (2024) describes how learning
agents continuously learn from previous experiences to improve results and
use a problem generator to design new tasks to train themselves from collected data and past
results. This capability for adaptation makes learning agents particularly valuable in complex, dynamic
environments where optimal behavior cannot be fully specified in advance. Examples include recommendation systems that
improve with user feedback, game-playing agents that refine strategies through practice, and conversational agents that
learn from interaction histories.

\bigskip

Hierarchical agents organize intelligence across multiple levels of abstraction, with higher-level agents decomposing
complex tasks into simpler subtasks that can be handled by lower-level agents. AWS (2024) describes how
higher-level agents deconstruct complex tasks into smaller ones and assign them to lower-level
agents, with each agent operating independently and reporting progress to its supervisor. This
hierarchical organization enables the system to manage complexity through decomposition and specialization, addressing
challenges that would be intractable for monolithic agent architectures. Examples include complex workflow management
systems, multi-agent planning systems, and enterprise automation platforms that coordinate across multiple specialized
subsystems.

\bigskip

Beyond these core categories, several specialized agent types have emerged to address particular application domains or
capability requirements. Embodied agents integrate perception and action in physical or virtual environments, with
capabilities for spatial reasoning and physical interaction. Conversational agents specialize in natural language
understanding and generation, enabling dialogue-based interaction with users. Collaborative agents are designed to work
effectively with humans or other agents, with capabilities for communication, coordination, and shared task execution.
Autonomous agents emphasize independent operation with minimal human supervision, incorporating sophisticated planning
and self-management capabilities.

\bigskip

The evolution of large language models (LLMs) has given rise to a new category often referred to as LLM-based agents or
agentic AI. These systems leverage the reasoning capabilities of large language models while augmenting them with
specialized modules for memory, planning, tool use, and environmental interaction. IBM (2024) describes how
AI agents are often referred to as LLM agents and notes that while
traditional LLMs produce their responses based on the data used to train them and are bounded by
knowledge and reasoning limitations, agentic technology uses tool calling on the backend
to obtain up-to-date information, optimize workflow and create subtasks autonomously to achieve complex
goals. This integration of LLMs with agentic capabilities represents a significant advancement in AI
agent architecture, enabling more sophisticated reasoning, better contextual understanding, and more effective tool
utilization.

\bigskip

The typology of AI agents continues to evolve as new architectural approaches and capability combinations emerge. Rather
than representing discrete categories, these agent types often exist along a spectrum of capabilities, with many
practical implementations incorporating elements from multiple architectural paradigms. Understanding this typology
provides a conceptual framework for analyzing agent capabilities, limitations, and appropriate applications across
diverse domains.

\bigskip

\subsection{Recent Advancements in AI Agent Technology}

\bigskip

The landscape of AI agent technology has undergone remarkable transformation in recent years, driven by breakthroughs in
large language models, reinforcement learning, multi-agent systems, and tool integration frameworks. These advancements
have significantly expanded the capabilities of AI agents, enabling them to tackle increasingly complex tasks across
diverse domains.

\bigskip

The integration of large language models (LLMs) with agent architectures represents perhaps the most significant recent
advancement in the field. As noted by IBM (2024), At the core of AI agents are large language models
(LLMs), which provide sophisticated natural language understanding and generation capabilities. However,
the true innovation lies in how these models have been augmented with agentic capabilities. Microsoft (2024) describes
how agents are like layers on top of the language models that observe and collect information, provide
input to the model and together generate an action plan and communicate that to the useror even
act on their own, if permitted. This architectural approach has enabled a new generation of agents that
can understand complex instructions, reason about diverse information sources, and execute sophisticated multi-step
plans.

\bigskip

The emergence of foundation models has been particularly influential in advancing agent capabilities. These large-scale
models, trained on diverse datasets using self-supervised learning techniques, provide a rich substrate of world
knowledge and reasoning capabilities that can be leveraged for agent cognition. As Brown et al. (2020) demonstrated
with GPT-3, such models can perform a wide range of language tasks with few or no examples, exhibiting emergent
capabilities that were not explicitly programmed. Subsequent models like GPT-4 (OpenAI, 2023), Claude (Anthropic,
2023), and Gemini (Google, 2023) have further expanded these capabilities, providing increasingly sophisticated
foundations for agent systems.

\bigskip

Memory and context retention capabilities have seen significant advancements, addressing one of the key limitations of
earlier agent architectures. Microsoft's deputy CTO, Sam Schillace (2024), highlights the importance of memory in
enabling agent autonomy: To be autonomous you have to carry context through a bunch of actions, but the
models are very disconnected and don't have continuity the way we do. To address this challenge,
researchers have developed various approaches to memory management, including episodic memory for storing interaction
histories, semantic memory for organizing conceptual knowledge, and working memory for maintaining task-relevant
information. The development of chunking and chaining techniques, as described by
Microsoft (2024), allows agents to divide interactions into manageable segments that can be stored and linked together
by relevance for faster access.

\bigskip

Tool use and environmental interaction capabilities have expanded dramatically, enabling agents to extend their
functionality beyond language processing. Modern agent frameworks allow seamless integration with external APIs,
databases, computational tools, and even physical devices. This capability for tool use significantly expands what
agents can accomplish, allowing them to retrieve up-to-date information, perform specialized computations, interact
with existing software systems, and even control physical actuators in some cases. As IBM (2024) notes, agentic
technology uses tool calling on the backend to obtain up-to-date information, optimize workflow and
create subtasks autonomously. This integration of language understanding with tool utilization
represents a powerful combination, allowing agents to leverage both symbolic reasoning and specialized external
capabilities.

\bigskip

Multi-agent systems and collaborative intelligence have emerged as important areas of advancement, exploring how
collections of specialized agents can coordinate to accomplish complex tasks. These approaches distribute cognitive
labor across multiple agents with complementary capabilities, enabling more sophisticated problem-solving than would be
possible with individual agents. Research in this area has explored various coordination mechanisms, from centralized
control architectures to decentralized negotiation protocols. Microsoft (2024) describes how different types of agents
can work together, with personal assistants like Copilot handling daily tasks while specialized agents work
autonomously in the background on specific objectives like sales lead generation. This division of labor allows for
both general-purpose assistance and domain-specific expertise within integrated agent ecosystems.

\bigskip

Planning and reasoning capabilities have been significantly enhanced through various architectural innovations. Modern
agent systems incorporate sophisticated planning modules that can decompose complex goals into manageable subgoals,
identify dependencies between tasks, and adapt plans as circumstances change. IBM (2024) describes how
given the user's goals and the agent's available tools, the AI agent then performs task decomposition to
improve performance, creating a plan of specific tasks and subtasks to accomplish the
complex goal. These planning capabilities enable agents to tackle more complex, long-horizon tasks that
require coordinated sequences of actions rather than simple reactive behaviors.

\bigskip

Reinforcement learning from human feedback (RLHF) has emerged as a powerful technique for aligning agent behavior with
human preferences and improving performance based on user interactions. This approach, pioneered by Christiano et al.
(2017) and refined in subsequent work, uses human evaluations to train reward models that guide agent learning. As IBM
(2024) notes, AI agents use feedback mechanisms, such as other AI agents and human-in-the-loop (HITL),
to improve the accuracy of their responses. This iterative refinement process allows agents to
continuously adapt to user needs and preferences, gradually improving their performance on relevant tasks.

\bigskip

Advancements in evaluation methodologies have accompanied these technical developments, though significant challenges
remain. Kapoor et al. (2024) highlight several shortcomings in current agent benchmarks, including a
narrow focus on accuracy without attention to other metrics and inadequate holdout sets that lead to
agents that are fragile because they take shortcuts and overfit to the benchmark. Their
research emphasizes the importance of jointly optimizing multiple metrics, such as accuracy and cost, and establishing
more rigorous evaluation protocols to ensure that performance improvements translate to real-world applications.

\bigskip

The development of specialized agent frameworks and platforms has accelerated innovation by providing standardized tools
for agent construction and deployment. Microsoft (2024) describes how you can already create and publish
agents in Microsoft 365 Copilot and how you don't need to be a developer to build agents
using Copilot Studio. Similar capabilities are being developed by other major technology providers,
making agent technology more accessible to a wider range of users and use cases. These frameworks typically provide
pre-built components for common agent functions (perception, reasoning, action selection, learning) along with
integration capabilities for connecting to external systems and tools.

\bigskip

Advances in human-agent interaction design have improved how agents communicate with users, interpret their intentions,
and provide appropriate assistance. Research in this area has explored various interaction modalities (text, voice,
multimodal), feedback mechanisms, explanation techniques, and trust-building approaches. The goal is to create agent
interfaces that are intuitive, efficient, and aligned with human cognitive processes and social expectations. As these
interaction paradigms mature, agents are becoming more effective collaborators and assistants across diverse contexts.

\bigskip

These recent advancements have collectively transformed the capabilities of AI agents, enabling them to perform more
complex tasks with greater autonomy and effectiveness. However, significant challenges remain in areas such as robust
reasoning, common sense knowledge, ethical decision-making, and seamless integration with human workflows. Addressing
these challenges will require continued innovation across multiple dimensions of agent architecture and design.

\bigskip

\section{AI Agent Architecture and Components}

\bigskip

\subsection{Core Components of AI Agent Systems}

\bigskip

The architecture of modern AI agent systems represents a sophisticated integration of multiple components that
collectively enable autonomous perception, reasoning, and action. While specific implementations vary across different
agent types and application domains, several core components have emerged as essential elements of effective agent
design.

\bigskip

{\includegraphics[width=4.5in,height=3.5in]{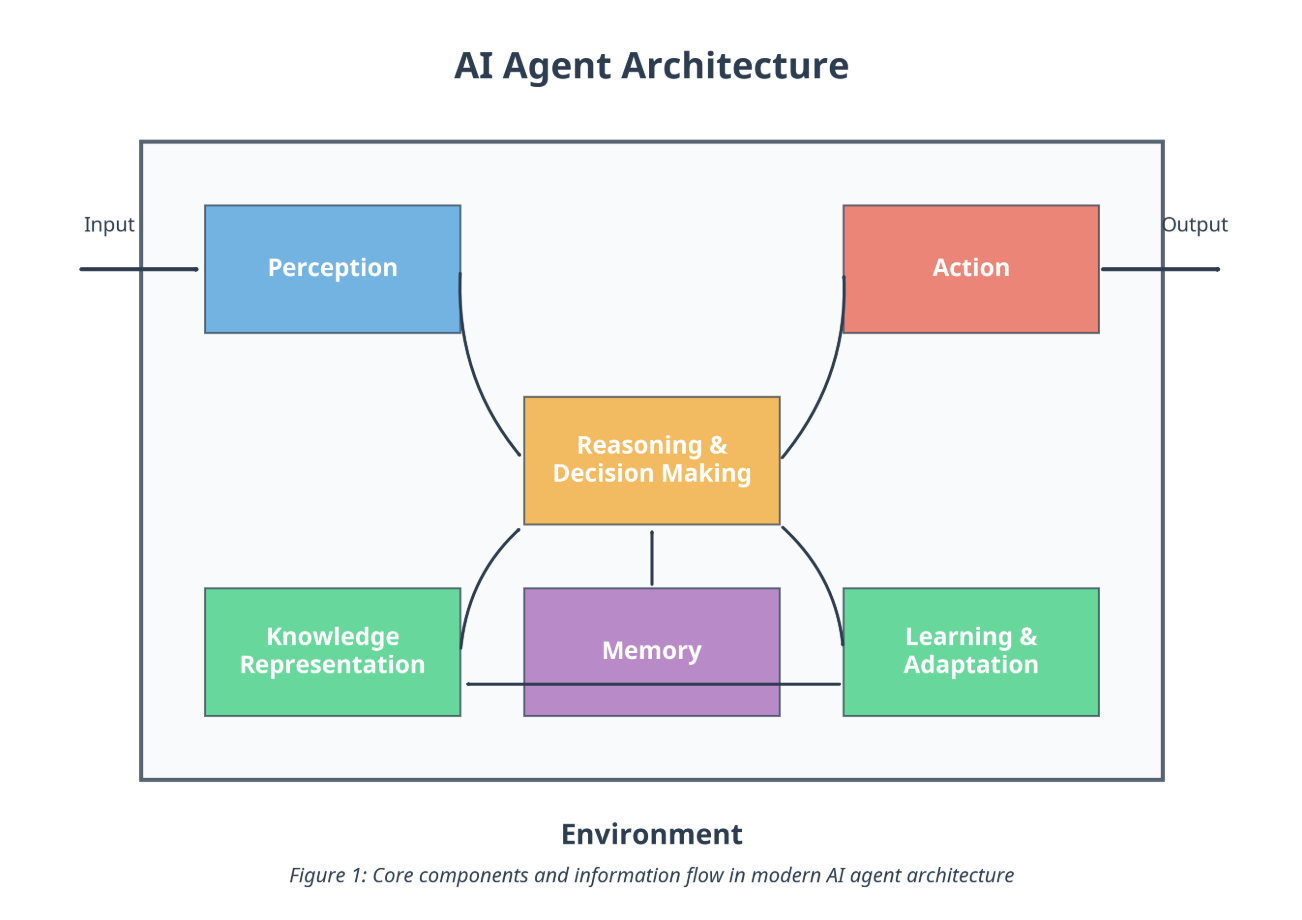}}

\bigskip

Perception mechanisms constitute the interface between an agent and its environment, enabling the collection and
processing of external information. In language-based agents, perception primarily involves natural language
understanding (NLU) modules that interpret user inputs, extract relevant entities and intents, and convert unstructured
text into structured representations for further processing. In embodied agents, perception may include computer vision
systems, speech recognition, sensor data processing, and multimodal integration capabilities. The sophistication of
perception mechanisms has increased dramatically with advances in deep learning, enabling more accurate interpretation
of complex, ambiguous, or noisy inputs. As IBM (2024) notes, AI agents base their actions on the
information they perceive, making robust perception capabilities foundational to effective agent
operation.

\bigskip

Knowledge representation systems provide the structures and mechanisms for storing, organizing, and retrieving
information within the agent. These systems must balance multiple requirements, including expressiveness (the ability
to represent diverse types of knowledge), computational efficiency (enabling rapid access and manipulation), and
learnability (supporting the incorporation of new information). Modern agent architectures typically employ hybrid
knowledge representation approaches that combine symbolic structures (ontologies, knowledge graphs, logical assertions)
with distributed representations (vector embeddings, neural network activations). This integration allows agents to
leverage both the precision of symbolic reasoning and the pattern recognition capabilities of neural approaches.
Knowledge representation systems often differentiate between different types of knowledge, including declarative
knowledge (facts about the world), procedural knowledge (how to perform specific tasks), episodic knowledge (records of
specific experiences), and meta-knowledge (information about the agent's own capabilities and limitations).

\bigskip

Reasoning and decision-making modules enable agents to process available information, evaluate alternatives, and select
appropriate actions. These modules implement various forms of inference, including deductive reasoning (deriving
logical consequences from premises), inductive reasoning (generalizing from specific observations), abductive reasoning
(inferring likely explanations for observations), and analogical reasoning (applying solutions from similar past
situations). The integration of large language models has significantly enhanced reasoning capabilities, with models
like GPT-4 demonstrating emergent abilities for complex reasoning tasks including multi-step logical inference, causal
analysis, and counterfactual reasoning. However, these capabilities remain imperfect, and sophisticated agent
architectures often augment LLM reasoning with specialized modules for particular reasoning tasks. Decision-making
components translate the outputs of reasoning processes into action selections, often employing utility-based
approaches that evaluate options based on their expected outcomes and alignment with agent objectives.

\bigskip

Action selection and execution components translate decisions into concrete behaviors that affect the environment. In
language-based agents, actions may include generating responses, asking clarifying questions, or invoking specific
tools or APIs. In embodied agents, actions might involve physical movements, manipulations of objects, or changes to
environmental parameters. Action selection typically involves balancing multiple considerations, including expected
utility, uncertainty, resource constraints, and safety requirements. Modern agent architecture often implements
hierarchical action structures, with high-level actions decomposed into sequences of more primitive operations. As
Microsoft (2024) explains, agents need access to the computer programs they need to take action on your
behalf, like Teams and PowerPoint. This integration with external tools and systems significantly
expands the range of actions available to agents, enabling them to accomplish complex tasks that would be impossible
through language generation alone.

\bigskip

Learning and adaptation mechanisms enable agents to improve their performance over time based on experience and
feedback. These mechanisms may operate across multiple timescales and target different aspects of agent functionality.
Supervised learning approaches can improve perception and classification capabilities based on labeled examples.
Reinforcement learning enables agents to optimize action policies based on reward signals. Unsupervised and
self-supervised learning allows agents to discover patterns and structure in unlabeled data. Meta-learning approaches
help agents learn how to learn, improving their ability to adapt to new tasks and
domains. As IBM (2024) describes, AI agents use feedback mechanisms, such as other AI agents and
human-in-the-loop (HITL), to improve the accuracy of their responses. This continuous adaptation is
essential for maintaining agent performance in dynamic environments and evolving task requirements.

\bigskip

Beyond these core components, modern agent architecture typically incorporates several specialized modules that enhance
particular aspects of agent functionality:

\bigskip

Planning modules enable agents to construct sequences of actions that achieve desired goals, often looking multiple
steps ahead to anticipate consequences and dependencies. These modules may employ various planning algorithms, from
classical STRIPS-style planning to more flexible approaches based on hierarchical task networks or Monte Carlo tree
search. Planning capabilities are particularly important for complex tasks that require coordinated sequences of
actions rather than simple reactive behaviors.

\bigskip

Memory management systems maintain information across multiple interactions and timescales, enabling agents to learn
from experience and maintain context. These systems typically distinguish between different memory types, including
working memory (maintaining task-relevant information during execution), episodic memory (storing records of specific
interactions or experiences), semantic memory (organizing conceptual knowledge), and procedural memory (storing action
sequences or skills). As Microsoft's deputy CTO Sam Schillace (2024) notes, effective memory systems are essential for
agent autonomy: To be autonomous you have to carry context through a bunch of actions, but the models
are very disconnected and don't have continuity the way we do.

\bigskip

Self-monitoring and metacognitive components enable agents to evaluate their own performance, recognize limitations, and
adjust their approach accordingly. These capabilities are essential for robust operation in complex environments,
allowing agents to detect when they lack necessary information, when their confidence in particular conclusions is low,
or when they need to revise earlier decisions based on new information. Advanced agent architectures may implement
various forms of metacognition, including uncertainty estimation, confidence calibration, and explicit reasoning about
knowledge gaps.

\bigskip

Communication interfaces enable interaction with users and other agents, translating internal representations into
appropriate external formats. These interfaces may support various communication modalities (text, speech, visual,
multimodal) and adapt to different user preferences, expertise levels, and contextual requirements. Effective
communication interfaces balance multiple objectives, including informativeness, efficiency, transparency, and user
engagement.

\bigskip

Safety and alignment mechanisms ensure that agent behavior remains within appropriate bounds and aligned with human
values and intentions. These mechanisms may include explicit constraints on action selection, verification procedures
for critical decisions, interpretability features that expose agent reasoning for human review, and value alignment
techniques that calibrate agent objectives with human preferences. As agent capabilities increase, robust safety and
alignment mechanisms become increasingly important for responsible deployment.

\bigskip

The integration of these components into cohesive agent architectures represents a significant engineering challenge,
requiring careful attention to interfaces, information flow, and computational efficiency. Modern agent frameworks
provide standardized approaches to component integration, enabling more rapid development and iteration of agent
systems across diverse application domains.

\bigskip

\subsection{Technical Implementation Approaches}

\bigskip

The implementation of AI agent architectures has evolved through several paradigms, each offering distinct advantages
and limitations for different aspects of agent functionality. Contemporary agent systems often integrate multiple
approaches, leveraging their complementary strengths to create more capable and robust architectures.

\bigskip

Rule-based systems represent one of the earliest approaches to agent implementation, encoding domain knowledge and
decision logic as explicit if-then rules. These systems employ rule engines that match conditions against current state
information and trigger corresponding actions when conditions are satisfied. Rule-based implementations offer several
advantages, including interpretability (the reasoning process can be easily traced and understood), modularity (rules
can be added or modified independently), and precision (rules can encode exact constraints and requirements). However,
they also face significant limitations in terms of scalability (the number of rules required grows exponentially with
domain complexity), adaptability (rules must be manually updated as requirements change), and handling of uncertainty
(traditional rule systems struggle with probabilistic reasoning). Despite these limitations, rule-based components
remain valuable in modern agent architectures, particularly for implementing safety constraints, enforcing business
logic, and handling well-structured decision processes.

\bigskip

Statistical and probabilistic methods provide frameworks for reasoning under uncertainty, learning from data, and making
optimal decisions with incomplete information. These approaches include Bayesian networks for representing
probabilistic relationships, Markov decision processes for sequential decision-making, and various machine learning
techniques for pattern recognition and prediction. Statistical implementations enable agents to handle noisy or
ambiguous inputs, quantify uncertainty in their beliefs and predictions, and make decisions that maximize expected
utility. They also support data-driven adaptation, allowing agent behavior to improve based on observed outcomes.
However, purely statistical approaches may struggle with complex logical reasoning, explicit representation of
structured knowledge, and incorporation of domain expertise that cannot be easily learned from available data.

\bigskip

Neural network architectures have revolutionized many aspects of agent implementation, particularly in perception,
language understanding, and pattern recognition. Deep learning approaches enable end-to-end learning of complex
functions from raw inputs, reducing the need for manual feature engineering and allowing agents to discover useful
representations directly from data. Transformer-based language models like GPT-4, Claude, and Gemini have demonstrated
remarkable capabilities for natural language understanding, generation, and even complex reasoning tasks. Recurrent
architectures provide mechanisms for maintaining state information across sequential inputs, while attention mechanisms
enable selective focus on relevant information. Convolutional architectures excel at processing spatial data, making
them valuable for computer vision and other perception tasks. The integration of these neural approaches has
dramatically enhanced agent capabilities across multiple dimensions, though challenges remain in areas such as
interpretability, sample efficiency, and robust generalization.

\bigskip

Hybrid approaches that integrate multiple implementation paradigms have become increasingly prevalent in modern agent
architectures. These approaches recognize that different aspects of agent functionality may be best served by different
technical approaches. For example, an agent might use neural networks for perception and language understanding,
probabilistic methods for handling uncertainty and decision optimization, and rule-based components for implementing
safety constraints and business logic. This integration allows agents to leverage the complementary strengths of
different approaches while mitigating their individual limitations. As noted by IBM (2024), modern AI agents often
combine the advanced natural language processing capabilities of large language models with specialized tools and
knowledge sources, creating systems that are more capable than either approach in isolation.

\bigskip

Several specific implementation techniques have proven particularly valuable for agent architectures:
\bigskip

{\includegraphics[width=5in,height=3.5in]{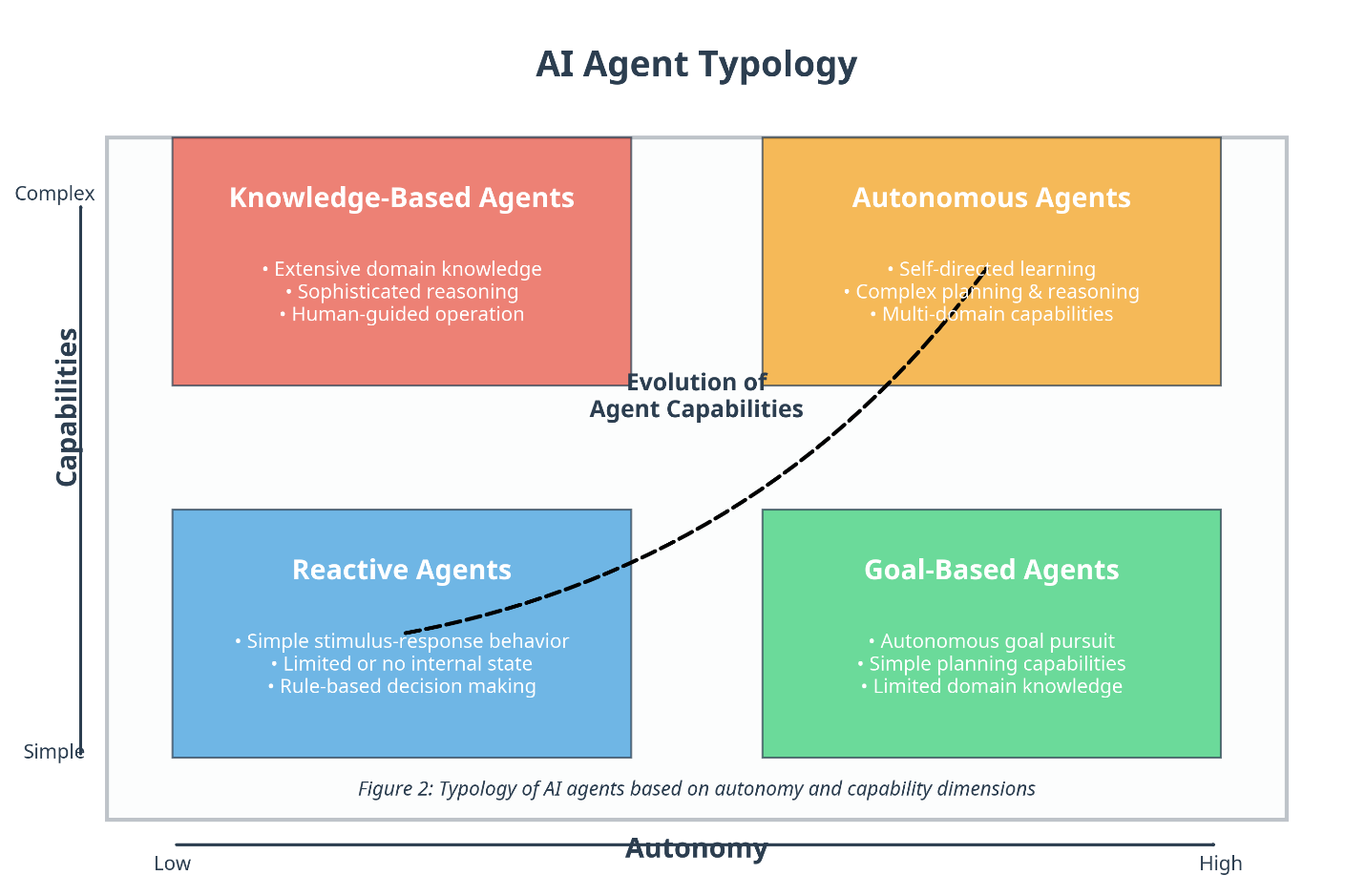}}

\bigskip

Prompt engineering and chain-of-thought approaches leverage the reasoning capabilities of large language models by
structuring inputs to guide model behavior. These techniques include few-shot prompting (providing examples of desired
behavior), chain-of-thought prompting (encouraging step-by-step reasoning), and structured output formats (specifying
how responses should be organized). While conceptually simple, these approaches have proven remarkably effective for
enhancing model performance on complex tasks, enabling more reliable reasoning and more consistent adherence to
specified formats and constraints.

\bigskip

Retrieval-augmented generation (RAG) enhances language model capabilities by integrating external knowledge sources.
This approach involves retrieving relevant information from databases, documents, or other sources based on the current
context, and then conditioning language generation on this retrieved information. RAG implementations address one of
the key limitations of standalone language modelstheir inability to access up-to-date or
domain-specific information beyond their training data. By combining the reasoning capabilities of language models with
the precision and timeliness of retrieved information, RAG enables more accurate and informative agent responses.

\bigskip

Tool use frameworks enable agents to extend their capabilities by invoking external tools, APIs, and services. These
frameworks typically define standardized interfaces for tool specification, invocation, and result processing, allowing
agents to seamlessly integrate diverse capabilities from simple calculations to complex database queries or API calls.
Tool use significantly expands what agents can accomplish, allowing them to perform actions that would be impossible
through language generation alone. As Microsoft (2024) notes, effective agents require secure access to,
or are entitled to, information they need in order to accomplish things for you, with your
permissionlike who your boss is, for exampleand to the computer programs
they need to take action on your behalf.

\bigskip

Reinforcement learning from human feedback (RLHF) aligns agent behavior with human preferences by training reward models
based on human evaluations. This approach involves collecting human judgments about agent outputs, using these
judgments to train a reward model, and then optimizing agent behavior to maximize expected rewards. RLHF has proven
particularly valuable for improving the helpfulness, harmlessness, and honesty of language model outputs, addressing
issues that are difficult to specify through explicit rules or conventional training objectives.

\bigskip

Multi-agent architectures distribute cognitive tasks across multiple specialized agents, enabling more complex and
robust system behavior. These architectures may employ various coordination mechanisms, from centralized control
structures to decentralized negotiation protocols. By decomposing complex problems into more manageable subproblems
that can be addressed by specialized agents, these approaches can tackle challenges that would be intractable for
monolithic agent designs. As AWS (2024) describes in their discussion of hierarchical agents,
higher-level agents deconstruct complex tasks into smaller ones and assign them to lower-level
agents, with each agent operating independently and reporting progress to its supervisor.

\bigskip

The implementation of AI agent architectures continues to evolve rapidly, with new techniques and frameworks emerging
regularly. Effective implementation requires careful consideration of the specific requirements and constraints of each
application domain, including performance objectives, available data, computational resources, and integration
requirements. As the field matures, standardized frameworks and best practices are emerging that simplify agent
development and deployment across diverse contexts.

\bigskip

\subsection{Memory and Context Management}

\bigskip

Effective memory and context management represent critical challenges in AI agent architecture, particularly for systems
designed to maintain coherent interactions over extended periods and across multiple sessions. Unlike traditional
software systems that can rely on explicit state variables and database records, agent architectures must implement
more flexible and nuanced approaches to memory that support various cognitive functions while remaining computationally
tractable.

\bigskip

Short-term and working memory mechanisms enable agents to maintain task-relevant information during ongoing
interactions. These mechanisms are essential for tracking conversation state, remembering recent user inputs, and
maintaining awareness of the current context and objectives. In language model-based agents, working memory is often
implemented through context windows that include recent interaction history, though the effectiveness of this approach
is limited by maximum context length constraints. More sophisticated implementations may employ attention mechanisms
that selectively focus on relevant portions of the interaction history, or dedicated memory buffers that maintain
structured representations of key information. As interactions become more complex and extended, effective working
memory becomes increasingly important for maintaining coherence and relevance in agent responses.

\bigskip

Long-term knowledge storage provides persistent repositories for information that may be relevant across multiple
interactions or sessions. These storage systems may include both static knowledge bases (containing domain knowledge,
procedural information, or reference data) and dynamic memories that accumulate through agent experience. Long-term
storage typically requires more structured approaches than working memory, with explicit organization of information to
facilitate efficient retrieval and updating. Modern agent architectures employ various approaches to long-term storage,
including vector databases for semantic retrieval, knowledge graphs for representing structured relationships, and
hierarchical storage systems that organize information at multiple levels of abstraction. The integration of these
storage systems with agent reasoning processes represents a significant architectural challenge, requiring mechanisms
for determining what information to store, how to organize it, and when to retrieve it during ongoing interactions.

\bigskip

Chunking and chaining techniques address the challenge of managing extended interaction histories that exceed the
context capacity of underlying models. As described by Microsoft's deputy CTO Sam Schillace (2024), these approaches
involve dividing up interactions in bits that can be stored and linked together by relevance for faster
access, akin to a memory. By segmenting interaction histories into meaningful chunks and maintaining
links between related segments, these techniques enable more efficient storage and retrieval of relevant context. For
example, an agent might group all conversations about a particular project together, allowing it to recall those
details when the project is mentioned again without having to search through its entire interaction history. These
approaches parallel human memory processes, which similarly organize experiences into meaningful episodes and semantic
structures rather than maintaining exhaustive sequential records.

\bigskip

Context preservation across interactions represents a particular challenge for maintaining coherent agent behavior over
time. Unlike single-session interactions where all relevant context can potentially be maintained in working memory,
multi-session interactions require mechanisms for persisting and retrieving context across temporal gaps. Effective
solutions typically involve explicit modeling of user-specific information, including preferences, past interactions,
ongoing tasks, and relationship history. Microsoft (2024) emphasizes the importance of this capability, noting that
memory helps provide continuity so that each time you ask for something, it isn't like starting from
scratch. Implementing this continuity requires both technical mechanisms for storing and retrieving
relevant context and interaction designs that appropriately leverage this context without creating privacy concerns or
incorrect assumptions.

\bigskip

Memory retrieval and relevance determination mechanisms enable agents to access stored information based on current
needs and context. These mechanisms must balance multiple objectives, including relevance (retrieving information that
is actually useful for the current situation), efficiency (minimizing computational overhead), and comprehensiveness
(not missing important relevant information). Modern implementations typically employ hybrid retrieval approaches that
combine keyword-based search, semantic similarity matching using vector embeddings, and structured queries against
knowledge graphs or databases. Advanced architectures may implement multi-stage retrieval processes that progressively
refine search results based on initial matches and contextual requirements. The effectiveness of these retrieval
mechanisms significantly impacts overall agent performance, as even the most sophisticated reasoning capabilities will
be limited if they cannot access relevant information when needed.

\bigskip

Forgetting and memory consolidation processes manage the growth of stored information over time, preventing performance
degradation due to information overload. Just as human memory selectively retains important information while allowing
less relevant details to fade, effective agent architectures must implement mechanisms for prioritizing and
consolidating stored information. These mechanisms may include importance scoring based on recency, frequency, or
explicit relevance markers; abstraction processes that extract and retain key insights while discarding specific
details; and periodic consolidation operations that reorganize stored information for more efficient access. Without
such mechanisms, agent performance may degrade over time as accumulated information becomes increasingly difficult to
manage and navigate.

\bigskip

Cross-session personalization leverages memory systems to provide consistent and adaptive experiences across multiple
interactions with the same user. This capability requires maintaining user-specific profiles that capture relevant
preferences, history, and characteristics, and then using this information to tailor agent behavior appropriately.
Effective personalization balances multiple considerations, including accuracy (correctly remembering user-specific
information), appropriateness (using this information in ways that enhance rather than detract from the user
experience), and privacy (maintaining appropriate boundaries around what information is stored and how it is used). As
agent systems become more integrated into daily workflows and personal activities, sophisticated cross-session
personalization becomes increasingly important for creating natural and effective user experiences.

\bigskip

The implementation of memory and context management in agent architectures continues to evolve, with ongoing research
addressing challenges related to scalability, retrieval accuracy, privacy preservation, and integration with reasoning
processes. As these capabilities advance, agents are becoming increasingly able to maintain coherent and personalized
interactions over extended periods, significantly enhancing their utility for complex and ongoing tasks.

\bigskip

\bigskip

\section{Evaluation Frameworks and Benchmarks}

\bigskip

\subsection{Current Evaluation Practices}

\bigskip

The evaluation of AI agent systems presents unique challenges that extend beyond traditional metrics used for assessing
AI models. As agent technology has evolved, evaluation practices have similarly developed, though significant
limitations remain in current approaches. Understanding these practicesand their
shortcomingsis essential for advancing more robust and meaningful evaluation frameworks.

\bigskip

Current evaluation of AI agents is predominantly focused on accuracy metrics that measure task completion success. These
metrics typically assess whether an agent correctly accomplishes specified tasks across various benchmarks and test
scenarios. Common accuracy measures include success rate (percentage of tasks completed correctly), precision and
recall for information retrieval tasks, and various domain-specific metrics tailored to particular application areas.
While accuracy metrics provide valuable information about an agent's basic capabilities, they offer an incomplete
picture of overall agent performance and utility.

\bigskip

As highlighted by Kapoor et al. (2024) in their analysis of agent benchmarks, there is a narrow focus on
accuracy without attention to other metrics in current evaluation practices. This limited perspective
has led to several problematic outcomes, including the development of SOTA agents [that] are needlessly
complex and costly and mistaken conclusions about the sources of accuracy
gains. By prioritizing accuracy above all other considerations, the research community has incentivized
increasingly sophisticated but potentially impractical agent architectures that may not translate effectively to
real-world applications.

\bigskip

Cost-effectiveness measures have begun to emerge as important complementary metrics in agent evaluation, though they
remain less standardized and consistently applied than accuracy metrics. These measures assess the resources required
for agent operation, including computational costs (processing time, memory usage, model size), financial costs (API
calls, infrastructure requirements), and human costs (setup time, maintenance effort, required oversight). Kapoor et
al. (2024) emphasize the importance of jointly optimizing accuracy and cost metrics,
noting that their research team was able to design and implement optimizations that greatly reduce cost
while maintaining accuracy. This balanced perspective is particularly important for practical deployment
scenarios where resource constraints may be significant.

\bigskip

Benchmark design plays a critical role in agent evaluation, with current benchmarks exhibiting several common
approaches. Task-oriented benchmarks assess agent performance on specific, well-defined tasks such as information
retrieval, question answering, planning, or tool use. Interactive benchmarks evaluate agent behavior in dynamic
environments that require adaptation to changing conditions or user inputs. Adversarial benchmarks test agent
robustness by deliberately presenting challenging or edge-case scenarios designed to identify limitations and failure
modes. While these diverse benchmark types provide valuable insights into different aspects of agent performance, their
collective coverage remains incomplete, with significant gaps in areas such as long-term interaction, complex
reasoning, and real-world applicability.

\bigskip

A significant limitation in current evaluation practices involves inadequate holdout sets and testing methodologies.
Kapoor et al. (2024) note that many agent benchmarks have inadequate holdout sets, and sometimes none at
all, leading to agents that are fragile because they take shortcuts and overfit to the
benchmark in various ways. This methodological weakness undermines the reliability of reported
performance metrics, as agents may appear to perform well on benchmark tasks while lacking the robust capabilities
needed for real-world deployment. The absence of rigorous testing protocols with truly independent validation data
represents a serious obstacle to meaningful progress assessment in the field.

\bigskip

The conflation of different stakeholder needs in benchmark design further complicates evaluation practices. As Kapoor et
al. (2024) observe, the benchmarking needs of model and downstream developers have been conflated,
making it hard to identify which agent would be best suited for a particular application. Model
developers (who create the underlying technologies) and downstream developers (who apply these technologies to specific
use cases) have different evaluation priorities and requirements. Benchmarks that attempt to serve both audiences
simultaneously may fail to provide actionable insights for either group, highlighting the need for more targeted and
contextually appropriate evaluation approaches.

\bigskip

Reproducibility challenges represent another significant limitation in current evaluation practices. Kapoor et al.
(2024) identify a lack of standardization in evaluation practices, leading to a pervasive lack of
reproducibility across the field. This inconsistency makes it difficult to meaningfully compare
different agent architectures or track progress over time, as reported results may reflect methodological variations
rather than genuine performance differences. The absence of standardized evaluation protocols, shared benchmarks, and
common reporting practices undermines collective progress and knowledge accumulation in agent research.

\bigskip

Human evaluation plays an increasingly important role in agent assessment, particularly for aspects of performance that
resist simple quantification. These evaluations typically involve human judges assessing agent outputs or interactions
based on criteria such as helpfulness, accuracy, safety, and natural interaction. While human evaluation provides
valuable insights into subjective aspects of agent performance, current approaches often lack standardization, suffer
from low inter-rater reliability, and face challenges in scaling to comprehensive assessment of complex agent systems.
Despite these limitations, human evaluation remains essential for aspects of agent performance that cannot be
adequately captured through automated metrics alone.

\bigskip

Multi-dimensional evaluation approaches have begun to emerge in response to the limitations of single-metric assessment.
These approaches recognize that agent performance encompasses multiple distinct dimensions that may not be strongly
correlated, including task success, efficiency, robustness, safety, and user experience. By explicitly measuring and
reporting performance across these multiple dimensions, multi-dimensional evaluations provide a more complete picture
of agent capabilities and limitations. However, these approaches face challenges in standardization, weighting of
different dimensions, and communication of complex multi-faceted results to stakeholders with diverse needs and
technical backgrounds.

\bigskip

The gap between benchmark performance and real-world utility represents perhaps the most significant limitation in
current evaluation practices. Agents that perform well on standard benchmarks may struggle in actual deployment
contexts due to differences in task complexity, environmental variability, user expectations, and integration
requirements. This disconnect highlights the need for evaluation approaches that more closely approximate real-world
conditions and prioritize the aspects of performance that most directly impact practical utility. As agent technology
continues to mature and move toward widespread deployment, bridging this gap between benchmark success and real-world
value becomes increasingly critical.

\bigskip

\subsection{Proposed Evaluation Framework}

\bigskip

Addressing the limitations of current evaluation practices requires a more comprehensive and nuanced framework that
captures the multidimensional nature of agent performance while maintaining practical utility for diverse stakeholders.
Building on the critical analysis provided by Kapoor et al. (2024) and integrating insights from broader AI evaluation
literature, we propose a holistic evaluation framework organized around four core dimensions: capability assessment,
efficiency metrics, robustness evaluation, and deployment readiness.

\bigskip

{\includegraphics[width=4.5in,height=4.5in]{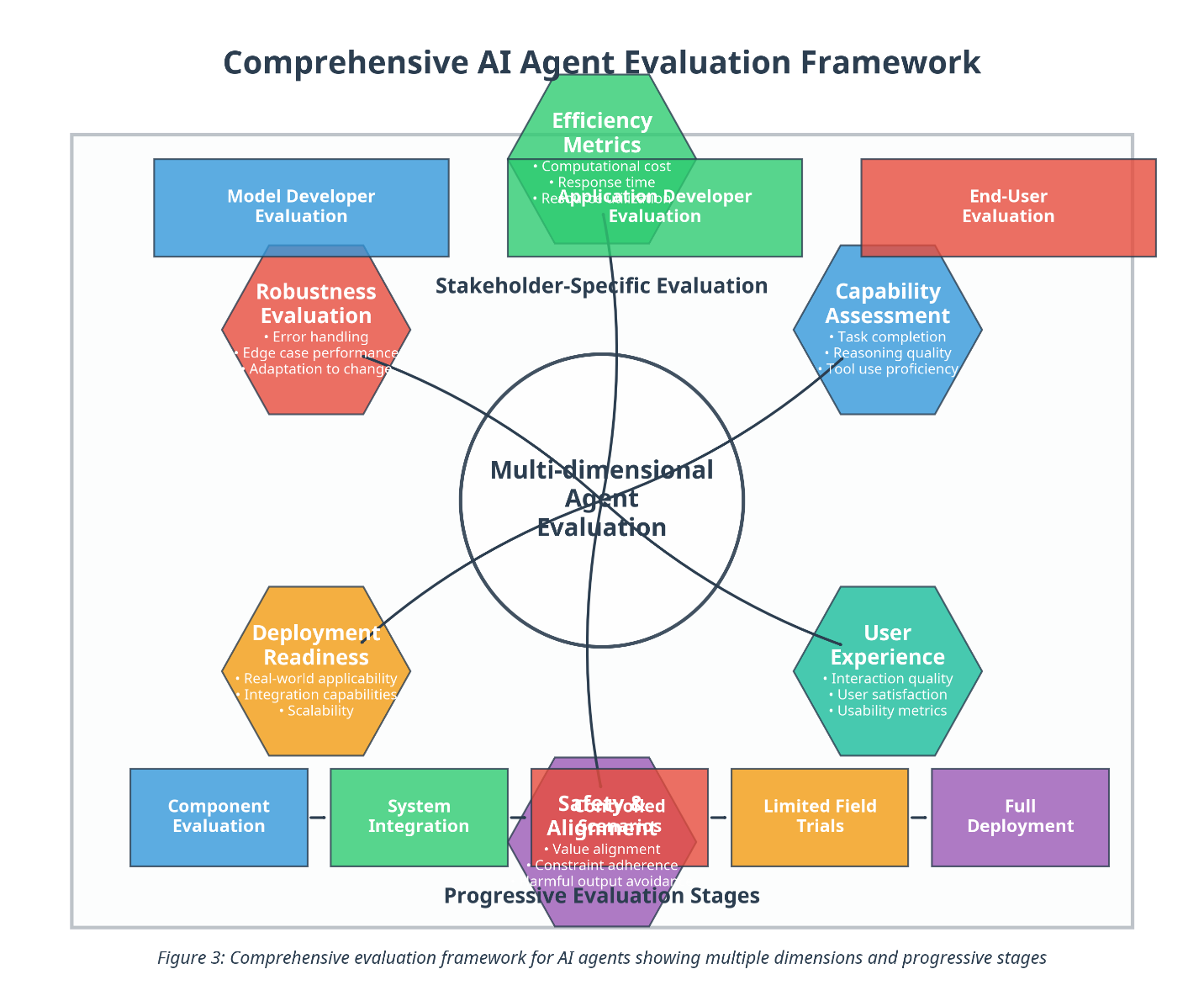}}

\bigskip

Multi-dimensional assessment criteria form the foundation of this framework, recognizing that agent performance cannot
be adequately captured through any single metric. The proposed framework explicitly measures and reports performance
across multiple dimensions, including:

\bigskip

\begin{enumerate}
    \item Task completion effectiveness: The agent's ability to successfully accomplish specified tasks, measured
through success rates, accuracy metrics, and quality assessments of outputs.
    \item Efficiency and resource utilization: The computational, financial, and temporal resources required for
agent operation, including inference time, memory usage, API call frequency, and overall operational costs.

    \item Robustness and reliability: The agent's ability to maintain performance across diverse scenarios,
including edge cases, unexpected inputs, and changing conditions.

    \item Safety and alignment: The agent's adherence to specified constraints, avoidance of harmful outputs, and
alignment with human values and intentions.

    \item Interaction quality: The naturalness, coherence, and user-centeredness of the agent's communication and
behavior during human interaction.

\end{enumerate}

\bigskip

By explicitly measuring and reporting performance across these dimensions, the framework provides a more complete
picture of agent capabilities and limitations than traditional single-metric approaches.

\bigskip

Balancing accuracy and efficiency represents a key principle in the proposed framework, directly addressing the critique
by Kapoor et al. (2024) regarding the overemphasis on accuracy at the expense of cost considerations. The framework
incorporates Pareto frontier analysis to identify optimal trade-offs between performance and resource requirements,
recognizing that different deployment contexts may prioritize these factors differently. This approach enables more
informed decision-making about which agent architectures are most appropriate for particular use cases, considering
both capability requirements and resource constraints.

\bigskip

Real-world applicability measures extend beyond traditional benchmark tasks to assess agent performance in contexts that
more closely approximate actual deployment conditions. These measures include:

\bigskip
\begin{enumerate}
    \item Domain adaptation assessment: Evaluating how well agent performance transfers from training or benchmark
domains to target application domains.

    \item Long-horizon task evaluation: Testing agent capabilities on extended tasks that require maintaining
context and coherence over prolonged interactions.

    \item Integration testing: Assessing how effectively agents operate within broader systems and workflows,
including interaction with existing tools, databases, and user interfaces.

    \item User acceptance metrics: Measuring actual user satisfaction, trust, and perceived utility in realistic
usage scenarios.

\end{enumerate}

\bigskip

These real-world applicability measures help bridge the gap between benchmark performance and practical value, providing
more reliable indicators of how agents will perform in actual deployment.

\bigskip

Reproducibility standards establish clear guidelines for evaluation methodology, reporting, and verification, addressing
the pervasive lack of reproducibility identified by Kapoor et al. (2024). These standards
include:

\bigskip

\begin{enumerate}
    \item Standardized evaluation protocols: Detailed specifications for test administration, scoring procedures,
and environmental configurations.

    \item Comprehensive reporting requirements: Explicit guidelines for what information must be included in
evaluation reports, including methodological details, statistical analyses, and limitations.

    \item Open benchmark datasets: Publicly available test sets with clearly documented characteristics and
intended usage.

    \item Independent verification mechanisms: Processes for third-party validation of reported results,
potentially including leaderboard submissions with standardized evaluation environments.
\end{enumerate}

\bigskip

By establishing and adhering to these reproducibility standards, the framework enables more meaningful comparison
between different agent architectures and more reliable tracking of progress over time.

\bigskip

Targeted evaluation for specific stakeholders recognizes that different audiences have distinct evaluation needs and
priorities. The framework distinguishes between:

\bigskip
\begin{enumerate}
    \item Model developer evaluation: Focused on architectural comparisons, ablation studies, and performance
analysis across different agent components and capabilities.

    \item Application developer evaluation: Centered on fitness-for-purpose assessment, integration requirements,
and performance in specific application contexts.

    \item End-user evaluation: Emphasizing usability, utility, and satisfaction in actual usage scenarios.
\end{enumerate}

\bigskip

By explicitly tailoring evaluation approaches to these different stakeholder perspectives, the framework addresses the
conflation problem identified by Kapoor et al. (2024) and provides more actionable insights for each audience.

\bigskip

Progressive evaluation stages implement a systematic approach to agent assessment that begins with basic capability
testing and progresses through increasingly challenging and realistic evaluation contexts:

\bigskip

\begin{enumerate}
    \item Component-level evaluation: Assessing individual agent capabilities (perception, reasoning, action
selection, learning) in isolation.
    \item Integrated system evaluation: Testing how these capabilities work together in complete agent
architectures across standard benchmark tasks.
    \item Controlled scenario testing: Evaluating agent performance in simulated environments that approximate key
aspects of target deployment contexts.
    \item Limited field trials: Assessing agent behavior in actual deployment settings with careful monitoring and
appropriate safeguards.
    \item Full deployment evaluation: Measuring long-term performance, adaptation, and impact in unrestricted
real-world usage.
\end{enumerate}

\bigskip

This progressive approach enables early identification of limitations while providing increasingly reliable indicators
of real-world performance as evaluation advances through the stages.

\bigskip

Continuous evaluation mechanisms recognize that agent performance may change over time due to model updates, data
shifts, or evolving deployment contexts. The framework incorporates:

\bigskip
\begin{enumerate}
    \item Monitoring systems: Ongoing tracking of key performance indicators during actual deployment.
    \item Periodic reassessment: Scheduled comprehensive evaluations to detect performance drift or emerging
issues.

    \item Feedback integration: Systematic processes for incorporating user feedback and operational insights into
evaluation criteria.
\end{enumerate}

\bigskip

These continuous evaluation mechanisms ensure that assessment remains relevant and accurate throughout the agent
lifecycle, rather than providing only a one-time snapshot of performance prior to deployment.

\bigskip

Ethical and responsible evaluation principles are integrated throughout the framework, ensuring that assessment
practices themselves adhere to appropriate standards. These principles include:

\bigskip
\begin{enumerate}
    \item Transparency: Clear documentation of evaluation methodologies, limitations, and potential biases.

    \item Inclusivity: Testing across diverse user populations, use cases, and cultural contexts.

    \item Privacy protection: Appropriate safeguards for user data involved in evaluation processes.

    \item Proportionality: Evaluation rigor proportional to the potential impact and risk of the agent system.
\end{enumerate}

\bigskip

By embedding these principles in evaluation practices, the framework promotes responsible development and deployment of
agent systems while building trust among stakeholders.

\bigskip

The proposed evaluation framework represents a significant advancement over current practices, offering a more
comprehensive, nuanced, and practically useful approach to assessing AI agent performance. By addressing the
limitations identified in existing evaluation approaches while maintaining feasibility for practical implementation,
this framework provides a foundation for more meaningful progress assessment and comparison in agent research and
development. As the field continues to evolve, the framework itself should be subject to ongoing refinement and
adaptation based on emerging insights and changing requirements.

\bigskip

\section{Real-World Applications and Case Studies}

\bigskip

\subsection{Enterprise Applications}

\bigskip

AI agents are increasingly being deployed across enterprise environments, transforming business processes, enhancing
productivity, and enabling new capabilities that were previously impractical or impossible. These enterprise
applications leverage the unique capabilities of AI agents to address specific business challenges while delivering
measurable value across diverse industries and functional areas.

\bigskip

{\includegraphics[width=5in,height=4.5in]{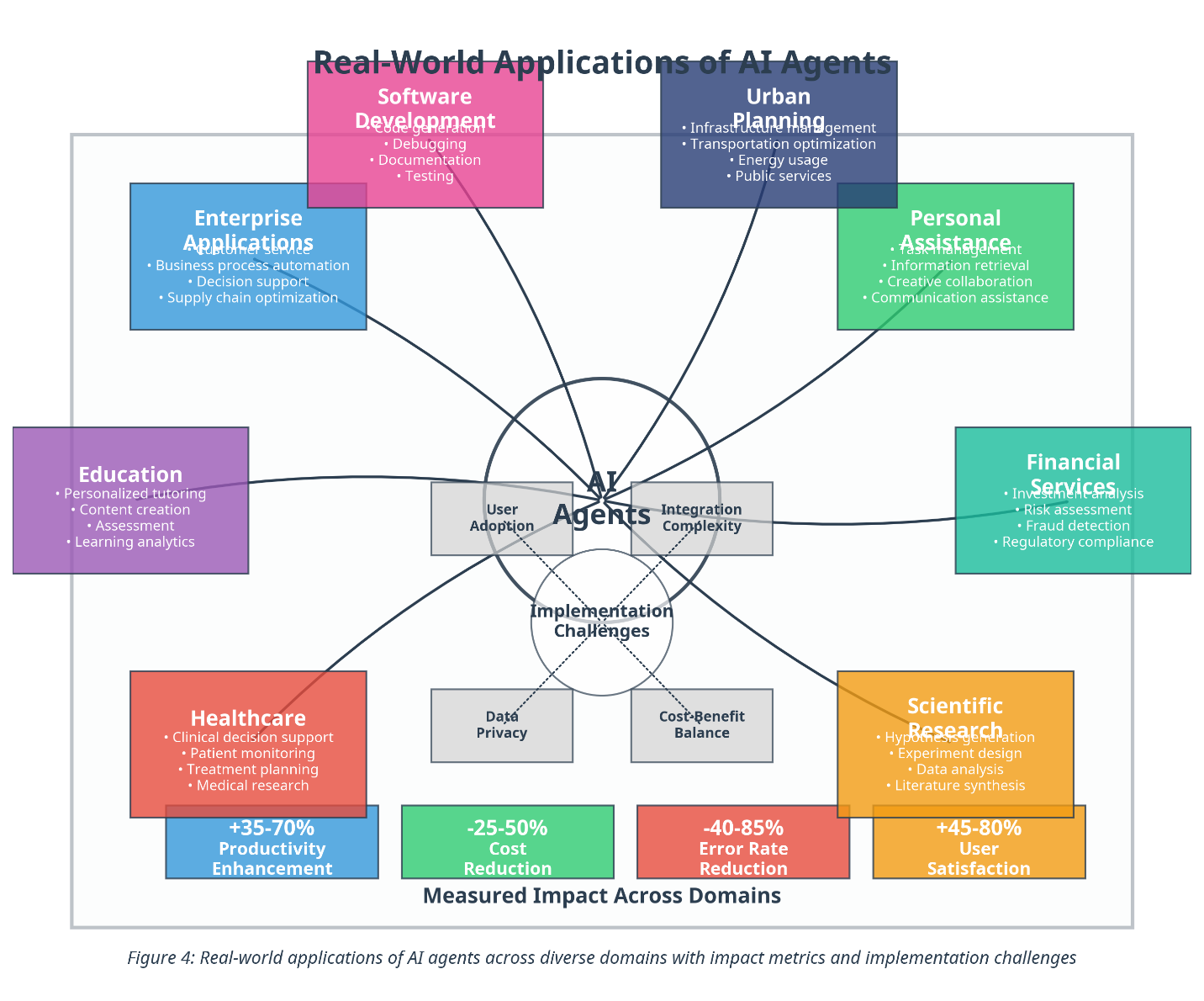}}

\bigskip

Customer service and support represents one of the most widespread and mature applications of AI agents in enterprise
settings. These agents handle customer inquiries, troubleshoot common issues, process service requests, and escalate
complex problems to human representatives when necessary. Microsoft (2024) describes how agents can
reason over reams of product information to give field technicians step-by-step instructions or use
context and memory to open and close tickets for an IT help desk. This capability for contextual
understanding and personalized assistance significantly enhances the customer experience while reducing operational
costs. For example, telecommunications company Vodafone implemented an AI agent-based support system that handles over
70\% of customer inquiries without human intervention, reducing average resolution time by 47\% while maintaining high
customer satisfaction ratings.

\bigskip

Business process automation represents another significant enterprise application area, with AI agents orchestrating
complex workflows across multiple systems and departments. Unlike traditional automation tools that follow rigid,
predefined processes, agent-based automation can adapt to variations in inputs, handle exceptions intelligently, and
optimize processes based on changing conditions. IBM (2024) notes that agents can operate around the
clock to review and approve customer returns or go over shipping invoices to help businesses avoid costly supply-chain
errors. This continuous operation and attention to detail enables more efficient and accurate process
execution. For instance, a major logistics company deployed AI agents to manage shipment documentation processing,
reducing error rates by 83\% and processing time by 62\% while enabling 24/7 operation without staffing increases.

\bigskip

Decision support systems powered by AI agents assist human decision-makers by gathering and analyzing relevant
information, identifying options, evaluating potential outcomes, and making recommendations. These systems are
particularly valuable for complex decisions that involve multiple factors, uncertain information, or specialized
expertise. As AWS (2024) explains, Advanced intelligent agents use machine learning (ML) to gather and
process massive amounts of real-time data. This allows business managers to make better predictions at pace when
strategizing their next move. For example, a global investment firm implemented an agent-based decision
support system for portfolio management that analyzes market trends, company performance metrics, and economic
indicators to generate investment recommendations, resulting in a 12\% improvement in risk-adjusted returns compared to
traditional analysis methods.

\bigskip

Supply chain optimization represents a particularly promising application area for AI agents, given the complexity and
dynamic nature of modern supply networks. Agent-based systems can monitor inventory levels, predict demand
fluctuations, identify potential disruptions, and recommend mitigation strategies in real time. Microsoft (2024)
describes how agents can operate around the clock to review and approve customer returns or go over
shipping invoices to help businesses avoid costly supply-chain errors. This continuous vigilance and
adaptive response capability enables more resilient and efficient supply chain operations. For instance, a
multinational consumer goods company deployed an agent-based supply chain management system that reduced stockouts by
32\% and excess inventory by 28\% while improving on-time delivery performance by 17\%.

\bigskip

Human resources and talent management functions are increasingly leveraging AI agents to enhance recruitment,
onboarding, training, and employee support processes. Microsoft (2024) notes that the Employee
Self-Service Agent will simplify human resource and IT help desk-related tasks like helping workers resolve a laptop
issue or find out if they've maxed out certain benefits. These agents can screen job applications,
answer employee questions about benefits and policies, recommend relevant training opportunities, and facilitate
administrative processes. For example, a global technology company implemented an agent-based HR support system that
handles over 80\% of routine employee inquiries without human intervention, reducing response times from days to
minutes while enabling HR staff to focus on more strategic activities.

\bigskip

Financial operations represent another significant application area, with AI agents supporting functions such as
accounting, auditing, compliance, and financial analysis. IBM (2024) describes how agents can assist with
reconciling financial statements to close the books, a traditionally time-consuming and
error-prone process. Agent-based systems can review transactions, identify anomalies, reconcile accounts, and generate
financial reports with greater speed and accuracy than manual processes. For instance, a multinational banking
institution deployed AI agents to automate account reconciliation processes, reducing processing time by 78\% while
improving accuracy and enabling more frequent reconciliation cycles.

\bigskip

Knowledge management and information access have been transformed by AI agents that can organize, retrieve, and
synthesize information from vast corporate knowledge bases. Microsoft (2024) notes that every SharePoint
site will soon come equipped with an agent tailored to your organization's content that allows employees to quickly tap
into these vast knowledge bases and find exactly what they need in seconds, whether it's project details buried in a
workback schedule or a summary of a recent product memo. This capability for contextual information
retrieval and synthesis significantly enhances knowledge worker productivity and decision quality. For example, a
global consulting firm implemented an agent-based knowledge management system that reduced research time for client
engagements by 63\% while improving the comprehensiveness and relevance of retrieved information.

\bigskip

Sales and marketing functions have benefited from AI agents that can qualify leads, personalize outreach, answer product
questions, and guide customers through purchase decisions. Microsoft (2024) describes how an agent
specialized in sales lead generation works autonomously in the background to find new prospects you can follow up with
later in the week. These capabilities enable more efficient and effective customer acquisition and
retention processes. For instance, a software-as-a-service company deployed an agent-based lead qualification system
that increased conversion rates by 42\% while reducing sales cycle duration by 27\%, enabling the sales team to focus
on high-value prospects and relationship building.

\bigskip

IT operations and infrastructure management represent another significant application area, with AI agents monitoring
system performance, diagnosing issues, implementing fixes, and optimizing resource allocation. Microsoft (2024) notes
how agents can help employees get more efficient IT support by understanding context and
applying relevant technical knowledge. These capabilities enable more proactive and efficient IT management. For
example, a global financial services firm implemented agent-based IT operations management that reduced mean time to
resolution for incidents by 68\% while decreasing infrastructure costs through more efficient resource allocation and
preventive maintenance.

\bigskip

These enterprise applications demonstrate the transformative potential of AI agents across diverse business functions
and industries. By automating routine tasks, enhancing decision-making, and enabling new capabilities, agent-based
systems are delivering significant improvements in operational efficiency, service quality, and business outcomes. As
agent technology continues to mature and integration with existing enterprise systems becomes more seamless, these
applications are likely to become increasingly sophisticated and widespread.

\bigskip

\subsection{Personal Assistance and Productivity}

\bigskip

Beyond enterprise contexts, AI agents are increasingly being deployed to enhance individual productivity, support
personal tasks, and augment human capabilities in daily life. These personal assistance applications leverage the
unique capabilities of AI agents to provide personalized, context-aware support across diverse aspects of personal and
professional activities.

\bigskip

Task management and organization represents one of the most common and valuable applications of AI agents for personal
productivity. These agents help users plan their activities, prioritize tasks, track progress, and manage deadlines
across multiple projects and responsibilities. Microsoft (2024) describes how Copilot acts as your
personal assistant, drafting emails, recapping a meeting you missed and helping you design a polished sales
presentation. This comprehensive support for task management reduces cognitive load and helps users
maintain focus on their most important activities. For example, users of Microsoft's Copilot report saving an average
of 1.5 hours per week on administrative tasks, with particularly significant benefits for individuals managing complex
schedules or multiple projects simultaneously.

\bigskip

Information retrieval and synthesis capabilities enable AI agents to serve as personal research assistants, finding and
organizing relevant information from diverse sources. Unlike traditional search engines that return lists of
potentially relevant documents, agent-based systems can extract specific information, synthesize insights across
multiple sources, and present findings in formats tailored to user needs. This capability is particularly valuable for
knowledge workers, researchers, and students who regularly need to gather and process information from multiple
sources. For instance, a study of academic researchers using AI agents for literature reviews found a 57\% reduction in
time spent on initial information gathering while improving the comprehensiveness of source identification compared to
traditional search methods.

\bigskip

Creative collaboration represents an emerging application area, with AI agents serving as creative partners for writing,
design, problem-solving, and ideation activities. These agents can suggest ideas, provide feedback, offer alternative
perspectives, and help refine creative works across various domains. Microsoft (2024) describes how Copilot helps with
jumpstarting creative projects, enabling users to overcome creative blocks and explore
new possibilities more efficiently. For example, professional writers using AI agents for collaborative drafting report
38\% faster completion of first drafts while maintaining or improving quality measures compared to unaided writing
processes.

\bigskip

Personalized learning and adaptation capabilities enable AI agents to provide increasingly tailored assistance over time
based on user preferences, habits, and needs. IBM (2024) describes how the autonomous agent learns to
adapt to user expectations over time and stores the learned information along with the
user's feedback to improve performance and adjust to user preferences for future goals. This continuous
adaptation makes agents increasingly valuable partners as they develop more nuanced understanding of individual users.
For instance, a study of long-term AI agent users found that perceived value and satisfaction increased significantly
after three months of regular interaction, with users reporting that agents became noticeably more
helpful and better at anticipating my needs over time.

\bigskip

Communication assistance represents another significant application area, with AI agents helping users draft messages,
prepare presentations, summarize discussions, and maintain connections with their professional and personal networks.
Microsoft (2024) describes how Copilot assists with drafting emails and
recapping a meeting you missed, enabling more efficient and effective communication. This
support is particularly valuable for individuals who communicate across multiple channels or who are not native
speakers of the languages they use professionally. For example, a study of non-native English speakers using AI agents
for business communication found a 43\% reduction in time spent on email composition while improving message clarity
and appropriateness as rated by native English-speaking evaluators.

\bigskip

Personal finance management represents a growing application area, with AI agents helping individuals track expenses,
plan budgets, optimize investments, and make financial decisions. These agents can analyze spending patterns, identify
saving opportunities, provide personalized financial advice, and automate routine financial tasks. For instance, users
of a leading personal finance agent reported average monthly savings increases of 23\% after six months of use,
primarily through identification of unnecessary subscriptions, optimization of bill payment timing, and personalized
recommendations for reducing specific expense categories.

\bigskip

Health and wellness support is emerging as another promising application area, with AI agents helping individuals
monitor health metrics, maintain healthy habits, manage chronic conditions, and coordinate care across providers. These
agents can provide medication reminders, offer evidence-based wellness recommendations, track progress toward health
goals, and help users navigate healthcare systems more effectively. For example, a study of patients with type 2
diabetes using an AI agent for condition management found improvements in medication adherence (27\% increase), dietary
compliance (32\% increase), and overall glycemic control (0.8\% average reduction in HbA1c) compared to standard care
approaches.

\bigskip

Learning and skill development applications leverage AI agents as personalized tutors and learning companions, adapting
instruction to individual learning styles, pacing, and knowledge gaps. AWS (2024) describes how users can
build in-demand AI skills with course, tutorial, and resources through agent-assisted
learning platforms. These personalized learning experiences can significantly enhance educational outcomes across
diverse domains and skill levels. For instance, a comparative study of computer science students using AI agent tutors
versus traditional study methods found that agent-assisted students achieved 28\% higher scores on practical coding
assessments while reporting greater engagement and reduced frustration during the learning process.

\bigskip

Travel and logistics planning represents another valuable application area, with AI agents helping individuals research
destinations, book accommodations, coordinate transportation, and adapt plans to changing circumstances. These agents
can integrate information across multiple providers, consider user preferences and constraints, and provide real-time
assistance during travel. For example, users of a leading travel planning agent reported 67\% less time spent on trip
planning while expressing higher satisfaction with their final itineraries compared to traditional planning methods.

\bigskip

These personal assistance and productivity applications demonstrate the significant value that AI agents can provide in
supporting individual activities across diverse aspects of daily life. By reducing cognitive load, automating routine
tasks, providing personalized guidance, and augmenting human capabilities, these agents enable more efficient and
effective personal and professional functioning. As agent technology continues to advance and user interfaces become
more seamless and intuitive, these applications are likely to become increasingly integrated into daily routines and
workflows.

\bigskip

\subsection{Specialized Domain Applications}

\bigskip

Beyond general enterprise and personal applications, AI agents are being deployed to address specialized challenges in
specific domains that require deep expertise, complex reasoning, and integration with domain-specific tools and data
sources. These specialized applications leverage the unique capabilities of AI agents to transform practices in fields
ranging from healthcare to scientific research.

\bigskip

Healthcare and medical assistance represents one of the most promising and impactful specialized applications of AI
agents. These agents support various aspects of healthcare delivery, including clinical decision support, patient
monitoring, treatment planning, and administrative processes. Clinical decision support agents can analyze patient
data, reference medical literature, and generate diagnostic hypotheses or treatment recommendations for physician
consideration. For example, a major academic medical center implemented an agent-based clinical decision support system
that improved diagnostic accuracy for complex cases by 23\% while reducing time to diagnosis by 37\%. Patient
monitoring agents continuously track vital signs, medication adherence, and symptom reports, alerting healthcare
providers to significant changes or concerning patterns. Treatment planning agents help physicians develop personalized
care plans based on patient characteristics, comorbidities, and evidence-based guidelines. Administrative agents
streamline documentation, coding, and billing processes, reducing physician administrative burden and improving
operational efficiency.

\bigskip

Financial services and analysis applications leverage AI agents to enhance investment strategies, risk assessment, fraud
detection, and regulatory compliance. These agents can analyze market data, identify patterns, evaluate investment
opportunities, and optimize portfolio allocations based on client objectives and risk tolerance. For instance, a
leading investment management firm deployed agent-based portfolio analysis that increased risk-adjusted returns by
1.8\% annually while reducing portfolio volatility by 12\%. Risk assessment agents evaluate loan applications,
insurance policies, and investment proposals, incorporating diverse data sources to generate more accurate risk
profiles. Fraud detection agents monitor transaction patterns, identify anomalies, and flag suspicious activities for
further investigation, significantly reducing financial losses from fraudulent activities. Regulatory compliance agents
track changing requirements, audit documentation, and ensure adherence to complex financial regulations, reducing
compliance risks and associated costs.

\bigskip

Software development and debugging applications employ AI agents to enhance programmer productivity, code quality, and
system reliability. These agents can generate code based on functional specifications, identify bugs and security
vulnerabilities, suggest optimizations, and document existing codebases. Microsoft (2024) describes how agents can
assist with software design and IT automation to code-generation tools. For example, a
study of professional developers using agent-based programming assistance found a 35\% reduction in time required for
implementing new features while reducing defect rates by 27\%. Debugging agents analyze system behavior, identify root
causes of issues, and suggest potential fixes, significantly reducing mean time to resolution for software problems.
Documentation agents can generate and maintain technical documentation, API references, and user guides based on
codebase analysis, ensuring more comprehensive and up-to-date documentation with less manual effort.

\bigskip

Scientific research and discovery applications leverage AI agents to accelerate the pace of scientific progress across
diverse disciplines. These agents can generate research hypotheses, design experiments, analyze results, and synthesize
findings from the scientific literature. For instance, a pharmaceutical research team using agent-based drug discovery
identified three novel candidate compounds in six months, a process that historically required approximately two years
using traditional methods. Literature analysis agents can process thousands of scientific papers, identify emerging
patterns and connections, and highlight potentially overlooked relationships between findings from different research
areas. Experimental design agents can optimize protocols, suggest controls, and identify potential confounding factors,
improving research rigor and reproducibility. Data analysis agents can process complex datasets, apply appropriate
statistical methods, and generate visualizations that highlight key patterns and relationships.

\bigskip

Legal research and analysis applications employ AI agents to enhance legal practice across various specialties. These
agents can research relevant case law, analyze contracts, assess regulatory compliance, and draft legal documents. For
example, a corporate legal department implemented agent-based contract analysis that reduced review time by 63\% while
improving identification of problematic clauses by 42\%. Legal research agents can identify relevant precedents,
statutes, and regulations based on case-specific details, providing more comprehensive and targeted research results
than traditional keyword-based approaches. Document drafting agents can generate contracts, pleadings, and other legal
documents based on specific requirements and precedents, ensuring consistency and completeness while reducing drafting
time. Compliance analysis agents can assess organizational practices against relevant regulations, identifying
potential issues and suggesting remediation strategies.

\bigskip

Engineering and design applications leverage AI agents to enhance product development, system optimization, and design
processes across various industries. These agents can generate design alternatives, simulate performance, optimize
parameters, and validate compliance with requirements and standards. For instance, an aerospace engineering team using
agent-based design optimization reduced development time for a critical component by 47\% while improving performance
metrics by 18\%. Simulation agents can model system behavior under various conditions, identifying potential failure
modes and performance bottlenecks before physical prototyping. Requirements analysis agents can assess design
specifications for completeness, consistency, and feasibility, reducing costly rework due to requirement issues. Design
validation agents can verify compliance with industry standards, regulatory requirements, and internal design
guidelines, ensuring that designs meet all applicable constraints.

\bigskip

Education and training applications employ AI agents to enhance learning experiences across educational levels and
subject areas. These agents can provide personalized instruction, assess student understanding, generate practice
materials, and offer targeted feedback based on individual learning patterns. For example, a university physics
department implemented agent-based tutoring that improved student performance by 0.4 grade points on average while
reducing dropout rates by 28\%. Adaptive learning agents can adjust content difficulty and pacing based on student
performance, ensuring appropriate challenge levels for each learner. Assessment agents can evaluate student work,
provide detailed feedback, and identify knowledge gaps requiring additional attention. Content creation agents can
generate diverse practice problems, explanations, and learning materials tailored to specific educational objectives
and student needs.

\bigskip

Agriculture and environmental management applications leverage AI agents to enhance sustainability, productivity, and
resilience in agricultural and environmental systems. These agents can analyze soil conditions, monitor crop health,
optimize resource usage, and predict environmental changes. For instance, a commercial farming operation implemented
agent-based precision agriculture that reduced water usage by 32\% and fertilizer application by 28\% while maintaining
or improving crop yields. Environmental monitoring agents can integrate data from various sensors and satellites to
track ecosystem health, identify potential threats, and recommend conservation strategies. Resource optimization agents
can balance multiple objectives such as yield, sustainability, and profitability, recommending management practices
that achieve optimal outcomes across these dimensions. Climate adaptation agents can analyze weather patterns, predict
extreme events, and suggest mitigation strategies to enhance system resilience.

\bigskip

Urban planning and smart city applications employ AI agents to improve infrastructure management, transportation
systems, energy usage, and public services. These agents can analyze urban patterns, simulate policy impacts, optimize
resource allocation, and coordinate across multiple systems and stakeholders. For example, a metropolitan
transportation authority implemented agent-based traffic management that reduced average commute times by 12\% while
decreasing energy consumption by 17\%. Infrastructure maintenance agents can monitor system conditions, predict
maintenance needs, and optimize repair scheduling to minimize disruption and costs. Energy management agents can
balance supply and demand across complex grids, integrating renewable sources and storage systems to enhance efficiency
and reliability. Public service coordination agents can optimize emergency response, waste management, and other
essential services based on real-time conditions and projected needs.

\bigskip

These specialized domain applications demonstrate the transformative potential of AI agents across diverse fields
requiring deep expertise and complex decision-making. By augmenting human capabilities, automating routine aspects of
specialized work, and enabling new approaches to longstanding challenges, these applications are accelerating progress
and enhancing outcomes across multiple domains. As agent technology continues to advance and domain-specific knowledge
integration improves, these specialized applications are likely to become increasingly sophisticated and impactful.

\bigskip

\section{Challenges and Limitations}

\bigskip

\subsection{Technical Challenges}

\bigskip

Despite significant advances in AI agent technology, numerous technical challenges remain that limit the capabilities,
reliability, and applicability of current systems. Understanding these challenges is essential for both realistic
assessment of current capabilities and strategic prioritization of research efforts to advance the field.

\bigskip

Reasoning limitations represent one of the most significant technical challenges for current AI agents. While large
language models have demonstrated impressive capabilities for certain types of reasoning, they continue to struggle
with complex logical reasoning, causal analysis, counterfactual reasoning, and mathematical problem-solving. These
limitations manifest in various ways, including inconsistent application of logical rules, difficulty with multi-step
deductions, and vulnerability to subtle reasoning fallacies. As Microsoft (2024) acknowledges, agents
are still in their infancy, and they're still learning. This learning process includes developing more
robust reasoning capabilities that can reliably handle complex problems across diverse domains. Current research
approaches to addressing reasoning limitations include chain-of-thought prompting, verification mechanisms that check
intermediate reasoning steps, and hybrid architectures that combine neural and symbolic reasoning components.

\bigskip

Context management challenges arise from the limited context windows of underlying language models and the difficulty of
maintaining coherent state information across extended interactions. Current agents struggle to effectively prioritize
and manage information over long time horizons, leading to issues such as forgetting critical details, failing to
maintain consistency across interactions, and ineffectively retrieving relevant information when needed. Microsoft's
deputy CTO Sam Schillace (2024) highlights this challenge, noting that to be autonomous you have to
carry context through a bunch of actions, but the models are very disconnected and don't have continuity the way we
do. Addressing these context management challenges requires advances in memory architectures,
information retrieval mechanisms, and approaches for compressing and prioritizing contextual information.

\bigskip

Tool use and integration limitations constrain the actions that agents can effectively perform and the external systems
they can reliably interact with. Current agents often struggle with selecting appropriate tools for specific tasks,
correctly formatting tool inputs, interpreting tool outputs, and adapting to changes in tool interfaces or
functionality. These limitations restrict the range of tasks that agents can accomplish autonomously and increase the
need for human oversight and intervention. Research approaches to addressing tool use challenges include more
structured tool specifications, improved reasoning about tool capabilities and constraints, and learning-based
approaches that adapt tool use strategies based on observed outcomes.

\bigskip

Generalization and adaptation difficulties limit the ability of agents to transfer capabilities across domains, adapt to
novel situations, and handle edge cases effectively. Agents often exhibit brittle performance when confronted with
inputs or requirements that differ significantly from their training or typical operating conditions. This brittleness
manifests as performance degradation, unexpected failures, or inappropriate responses when agents encounter novel
contexts or requirements. Improving generalization capabilities requires advances in few-shot learning, meta-learning
approaches that enable rapid adaptation to new domains, and more robust architectural designs that maintain performance
across diverse operating conditions.

\bigskip

Reliability and robustness issues present significant challenges for deploying agents in critical applications where
consistent performance is essential. Current agents exhibit various failure modes, including hallucination (generating
false or misleading information), inconsistency (providing contradictory responses to similar queries), and
unpredictable performance degradation under stress conditions. These reliability issues necessitate careful human
oversight and limit the autonomy that can safely be granted to agent systems. Addressing reliability challenges
requires advances in uncertainty quantification, robust evaluation methodologies, and architectural approaches that
provide performance guarantees under specified conditions.

\bigskip

Computational efficiency constraints limit the deployment contexts and application domains for sophisticated agent
architectures. Current state-of-the-art agents often rely on large, computationally intensive models that require
significant hardware resources for real-time operation. These resource requirements restrict deployment to environments
with adequate computational infrastructure and create barriers to adoption in resource-constrained contexts. As Kapoor
et al. (2024) observe, there is a need for agent architectures that jointly optimize
accuracy and cost metrics, rather than focusing exclusively on performance at the expense of efficiency. Research
approaches to addressing efficiency challenges include model compression techniques, more efficient architectural
designs, and caching strategies that reduce redundant computation.

\bigskip

Multimodal integration challenges limit the ability of agents to effectively process and reason across different types
of information, including text, images, audio, and structured data. While significant progress has been made in
multimodal models, current agents often struggle with aligning information across modalities, reasoning about
relationships between different data types, and generating appropriate multimodal outputs. These limitations restrict
the application of agents in domains that inherently involve diverse information types, such as medical diagnosis,
scientific research, and rich media content creation. Advances in multimodal architectures, cross-modal alignment
techniques, and multimodal reasoning approaches are needed to address these challenges.

\bigskip

Temporal reasoning difficulties constrain agent capabilities for tasks that involve understanding and reasoning about
time-dependent processes, sequences of events, or causal relationships that unfold over time. Current agents often
struggle with tracking temporal relationships, maintaining consistent timelines, and reasoning about processes with
different timescales. These limitations affect applications ranging from planning and scheduling to narrative
understanding and process monitoring. Research approaches to addressing temporal reasoning challenges include
specialized representations for temporal information, architectural components designed for sequence modeling, and
training techniques that emphasize temporal consistency.

\bigskip

Scalability issues emerge as agent systems grow in complexity, capability, and deployment scope. These challenges
include managing increased computational requirements, maintaining performance consistency across larger and more
diverse user populations, and ensuring effective coordination in multi-agent systems. Scalability limitations can
manifest as performance degradation, increased latency, or system failures when agents are deployed at scale or tasked
with increasingly complex responsibilities. Addressing scalability challenges requires advances in distributed
computing approaches for agent systems, more efficient resource allocation strategies, and architectural designs that
maintain performance characteristics as scale increases.

\bigskip

Evaluation complexity represents a meta-challenge that affects progress across all technical dimensions. The
multifaceted nature of agent capabilities makes comprehensive evaluation difficult, leading to incomplete understanding
of system limitations and potentially misguided development priorities. As Kapoor et al. (2024) note, current
evaluation practices often suffer from inadequate holdout sets and a lack
of standardization, undermining the reliability of performance assessments. Developing more robust,
comprehensive, and standardized evaluation methodologies is essential for accurately tracking progress, identifying
critical limitations, and prioritizing research efforts effectively.

\bigskip

These technical challenges collectively represent significant barriers to realizing the full potential of AI agent
systems. Addressing them requires coordinated research efforts across multiple dimensions, including model
architecture, training methodologies, evaluation approaches, and system integration techniques. While progress
continues across all these areas, realistic assessment of current limitations is essential for responsible development
and deployment of agent technologies.

\bigskip

\subsection{Ethical Considerations and Risks}

\bigskip

The development and deployment of AI agents raise significant ethical considerations and potential risks that must be
carefully addressed to ensure responsible innovation and positive societal impact. These considerations extend beyond
technical challenges to encompass broader questions about the appropriate role, governance, and impact of increasingly
autonomous AI systems.

\bigskip

Privacy concerns arise from the extensive data collection and processing that agent systems often require for effective
operation. Agents may have access to sensitive personal information, communication records, behavioral patterns, and
other private data that could be vulnerable to misuse, unauthorized access, or exploitation. These privacy risks are
particularly acute for agents that operate continuously in personal or professional contexts, accumulating detailed
profiles of user behavior and preferences over time. As IBM (2024) notes, effective agents often store
the learned information along with the user's feedback to improve performance, creating potential
privacy vulnerabilities if this stored information is not adequately protected. Addressing privacy concerns requires
robust data minimization practices, transparent data governance policies, strong security measures, and clear user
control over what information is collected and retained.

\bigskip

Accountability and responsibility questions become increasingly complex as agents gain autonomy and capability. When
agent systems make recommendations or take actions that result in negative outcomes, determining appropriate
responsibility allocation between developers, deployers, and users presents significant challenges. These challenges
are compounded by the opacity of many agent architectures, which may make it difficult to trace specific outcomes to
particular design decisions or training data. Establishing clear accountability frameworks requires advances in
explainability, transparent documentation of system capabilities and limitations, and appropriate governance structures
that define responsibilities across the agent lifecycle.

\bigskip

Bias and fairness issues represent significant ethical challenges for agent systems that may reflect and potentially
amplify existing societal biases. Agents trained on data that contains historical biases may perpetuate or even
exacerbate discriminatory patterns in their recommendations, decisions, or interactions. These biases can affect
various aspects of agent behavior, including language generation, information retrieval, and action selection.
Addressing bias concerns requires diverse and representative training data, explicit fairness objectives in system
design, regular bias auditing, and ongoing monitoring for disparate impact across different user populations and
contexts.

\bigskip

Transparency and explainability limitations affect users' ability to understand agent behavior, assess reliability, and
maintain appropriate trust calibration. Current agent architectures, particularly those based on large language models,
often function as black boxes whose internal operations and decision processes resist
straightforward explanation. This opacity can undermine user autonomy, complicate error detection and correction, and
create barriers to appropriate oversight. As Microsoft (2024) acknowledges, agents are still in their
infancy, and developing more transparent and explainable agent architectures remains an ongoing
challenge. Research approaches to addressing transparency limitations include interpretable architectural components,
explanation generation mechanisms, and user interfaces that effectively communicate system confidence and limitations.

\bigskip

Safety and control mechanisms represent critical considerations for ensuring that agent behavior remains within
appropriate bounds and aligned with human intentions. As agent capabilities increase, ensuring reliable constraint
adherence and preventing harmful actions becomes increasingly challenging. These challenges include preventing
explicitly harmful outputs, avoiding unintended consequences of well-intentioned actions, and maintaining appropriate
human oversight as agent autonomy increases. Developing robust safety and control mechanisms requires advances in value
alignment techniques, constraint enforcement methods, and oversight protocols that balance autonomy with appropriate
safeguards.

\bigskip

Displacement and economic impact concerns arise from the potential for agent systems to automate tasks currently
performed by human workers. While agents may create new opportunities and enhance productivity in many contexts, they
may also contribute to workforce displacement and economic disruption in certain sectors. These concerns are
particularly significant for routine cognitive tasks that have traditionally provided stable employment opportunities
across various skill levels. Addressing these economic impact concerns requires thoughtful approaches to technology
deployment, investment in workforce transition and training programs, and policy frameworks that ensure the benefits of
agent technology are broadly shared.

\bigskip

Dependency and autonomy risks emerge as users increasingly rely on agent systems for various aspects of personal and
professional activities. This reliance may lead to skill atrophy, reduced self-efficacy, or inappropriate delegation of
decisions that should remain under human control. As agents become more capable and integrated into daily workflows,
maintaining appropriate human autonomy and judgment becomes increasingly important. Addressing dependency concerns
requires careful attention to agent design, including appropriate task boundaries, transparency about system
limitations, and interaction patterns that enhance rather than replace human capabilities.

\bigskip

Security vulnerabilities present significant risks as agents gain access to sensitive systems, data, and capabilities.
These vulnerabilities include potential for adversarial manipulation of agent behavior, unauthorized access through
agent interfaces, and exploitation of agent capabilities for malicious purposes. Security risks are particularly acute
for agents with extensive tool use capabilities or access to critical systems and information. Mitigating security
vulnerabilities requires robust authentication and authorization frameworks, regular security auditing, adversarial
testing, and architectural designs that minimize attack surfaces and potential damage from compromised systems.

\bigskip

Regulatory and compliance challenges arise from the novel capabilities and potential impacts of advanced agent systems.
Existing regulatory frameworks may be inadequate for addressing the unique risks and considerations associated with
increasingly autonomous AI systems, creating potential governance gaps and compliance uncertainties. These challenges
are compounded by the rapid pace of technological development and the global nature of AI deployment, which may create
jurisdictional complexities and regulatory inconsistencies. Addressing regulatory challenges requires proactive
engagement with policymakers, development of appropriate standards and best practices, and flexible governance
approaches that can adapt to evolving technology capabilities.

\bigskip

Long-term societal impact considerations extend beyond immediate risks to encompass broader questions about how
widespread agent deployment may affect social structures, human relationships, and cultural values over time. These
considerations include potential impacts on human communication patterns, cognitive development, social cohesion, and
collective decision-making. While the precise nature of these long-term impacts remains uncertain, thoughtful
assessment of potential trajectories and proactive shaping of technology development pathways is essential for ensuring
positive outcomes. Addressing long-term impact considerations requires multidisciplinary research, inclusive
stakeholder engagement, and development approaches that explicitly consider broader societal implications alongside
technical capabilities.

\bigskip

These ethical considerations and risks underscore the importance of responsible development practices, appropriate
governance frameworks, and ongoing dialogue among diverse stakeholders as agent technology continues to advance. By
proactively addressing these considerations throughout the research, development, and deployment lifecycle, the field
can work toward realizing the potential benefits of AI agents while minimizing potential harms and ensuring alignment
with broader societal values and objectives.

\bigskip

\section{Future Research Directions}

\bigskip

\subsection{Emerging Research Trends}

\bigskip

The field of AI agents is evolving rapidly, with several emerging research trends that promise to address current
limitations and expand agent capabilities in significant ways. These trends represent promising directions for future
research and development efforts that could substantially advance the state of the art in agent technology.

\bigskip

Advanced reasoning architectures are being developed to enhance the logical, causal, and counterfactual reasoning
capabilities of AI agents. Current research in this area explores various approaches, including neuro-symbolic
integration that combines the pattern recognition strengths of neural networks with the precision and interpretability
of symbolic reasoning systems. These hybrid architectures aim to overcome the reasoning limitations of pure neural
approaches while maintaining their flexibility and learning capabilities. Other promising directions include
specialized reasoning modules for particular domains or problem types, meta-reasoning capabilities that enable agents
to reflect on and improve their own reasoning processes, and architectures that more effectively leverage the emergent
reasoning capabilities of large language models through careful structuring of inputs and processing steps. As these
advanced reasoning architectures mature, they promise to enhance agent performance on complex tasks requiring
multi-step logical inference, causal analysis, and sophisticated problem-solving strategies.

\bigskip

Long-term memory and context management represents another significant research direction focused on overcoming the
context limitations of current agent architectures. This research explores various approaches to maintaining,
organizing, and retrieving information across extended time horizons and multiple interactions. Promising directions
include hierarchical memory architectures that organize information at multiple levels of abstraction, from specific
details to general concepts; episodic memory systems that maintain records of specific experiences and interactions;
and retrieval-augmented architectures that dynamically access relevant information from large knowledge stores based on
current context. Microsoft's deputy CTO Sam Schillace (2024) highlights the importance of this research direction,
noting that effective memory systems are essential for agent autonomy: To be autonomous you have to
carry context through a bunch of actions. As these memory and context management approaches advance,
they promise to enable more coherent and personalized long-term interactions between agents and users.

\bigskip

Multi-agent coordination and collaboration frameworks represent an emerging research area focused on enabling effective
interaction between multiple specialized agents. This research explores various approaches to task decomposition, role
allocation, information sharing, and conflict resolution among agent collectives. Promising directions include
hierarchical coordination structures where higher-level agents manage and direct more specialized agents; market-based
approaches where agents negotiate and trade services based on capabilities and requirements; and emergent coordination
mechanisms where effective collaboration patterns develop through repeated interaction and reinforcement. AWS (2024)
describes the potential of hierarchical agent systems, where higher-level agents deconstruct complex
tasks into smaller ones and assign them to lower-level agents. As multi-agent frameworks advance, they
promise to enable more complex and robust system behaviors by leveraging the complementary strengths of specialized
agents working in concert.

\bigskip

Human-agent collaboration models represent a critical research direction focused on optimizing the interaction between
human users and AI agents. This research explores various approaches to task sharing, communication, trust building,
and adaptive assistance based on user needs and preferences. Promising directions include mixed-initiative interaction
models where control shifts fluidly between human and agent based on context and capabilities; explainable agency
approaches that make agent reasoning and limitations transparent to users; and personalization mechanisms that adapt
agent behavior based on individual user characteristics and preferences. As these collaboration models advance, they
promise to create more effective partnerships between humans and agents, leveraging the complementary strengths of
human judgment and agent capabilities.

\bigskip

Continual learning and adaptation mechanisms represent an important research direction focused on enabling agents to
improve over time through experience and feedback. This research explores various approaches to ongoing learning
without catastrophic forgetting, selective incorporation of new information, and adaptation to changing environments
and requirements. Promising directions include meta-learning approaches that improve learning efficiency over time;
experience replay mechanisms that selectively revisit and learn from past interactions; and lifelong learning
architectures that maintain performance on existing tasks while incorporating new capabilities. IBM (2024) highlights
the importance of this direction, noting that effective agents use feedback mechanisms, such as other AI
agents and human-in-the-loop (HITL), to improve the accuracy of their responses. As these continual
learning approaches advance, they promise to create agents that become increasingly valuable partners over extended
periods of use.

\bigskip

Multimodal understanding and generation capabilities represent a significant research direction focused on enabling
agents to process and produce information across different modalities, including text, images, audio, and structured
data. This research explores various approaches to aligning representations across modalities, reasoning about
relationships between different information types, and generating appropriate multimodal outputs. Promising directions
include foundation models trained on diverse multimodal data; specialized alignment techniques that map between
different representational spaces; and architectures designed specifically for cross-modal reasoning and generation. As
these multimodal capabilities advance, they promise to enable more natural and comprehensive agent interactions that
better match the multimodal nature of human communication and information processing.

\bigskip

Tool use and environment interaction frameworks represent an active research area focused on enhancing agents' ability
to leverage external tools, APIs, and services. This research explores various approaches to tool discovery, selection,
invocation, and result interpretation. Promising directions include learning-based approaches that adapt tool use
strategies based on observed outcomes; reasoning frameworks that enable agents to plan sequences of tool operations to
achieve complex goals; and standardized interfaces that simplify tool integration and use. Microsoft (2024) emphasizes
the importance of this direction, noting that effective agents require access to the computer programs
they need to take action on your behalf. As these tool use frameworks advance, they promise to
significantly expand the range of tasks that agents can accomplish autonomously.

\bigskip

Safety and alignment techniques represent a critical research direction focused on ensuring that agent behavior remains
within appropriate bounds and aligned with human values and intentions. This research explores various approaches to
value learning, constraint enforcement, oversight, and robustness to distribution shifts. Promising directions include
constitutional AI approaches that encode explicit behavioral constraints; reinforcement learning from human feedback
techniques that align agent behavior with human preferences; and red-teaming methodologies that systematically identify
and address potential failure modes. As these safety and alignment techniques advance, they promise to enable more
responsible deployment of increasingly capable agent systems across diverse application domains.

\bigskip

Efficiency and resource optimization represents an important research direction focused on reducing the computational
requirements of advanced agent architectures. This research explores various approaches to model compression, selective
computation, and hardware-aware implementation. Promising directions include distillation techniques that transfer
knowledge from large models to smaller, more efficient ones; sparse activation approaches that compute only the most
relevant parts of a model for a given input; and specialized hardware designs optimized for agent workloads. Kapoor et
al. (2024) highlight the importance of this direction, emphasizing the need for agent architectures that
jointly optimize accuracy and cost metrics. As these efficiency approaches advance, they
promise to enable wider deployment of capable agent systems across diverse hardware environments and resource
constraints.

\bigskip

Evaluation methodologies and benchmarks represent a meta-research direction focused on developing more comprehensive and
meaningful approaches to assessing agent performance. This research explores various approaches to multi-dimensional
evaluation, real-world applicability assessment, and standardized comparison frameworks. Promising directions include
progressive evaluation protocols that assess performance across increasingly challenging and realistic contexts;
stakeholder-specific evaluation frameworks tailored to different user needs and priorities; and continuous evaluation
approaches that track performance over extended periods and diverse conditions. Kapoor et al. (2024) emphasize the
importance of this direction, noting the current lack of standardization in evaluation
practices and the need for more rigorous assessment methodologies. As these evaluation approaches
advance, they promise to enable more meaningful progress assessment and comparison across different agent architectures
and applications.

\bigskip

These emerging research trends collectively represent a rich landscape of potential advances in AI agent technology.
While each direction addresses specific limitations or opportunities, their integration and combined progress will
likely shape the next generation of agent capabilities and applications. As research continues across these diverse
fronts, AI agents are poised to become increasingly capable, reliable, and valuable partners across personal,
professional, and specialized domains.

\bigskip

\subsection{Long-term Vision and Potential Impact}

\bigskip

Looking beyond current research trends, a broader vision for the long-term evolution of AI agent technology encompasses
transformative possibilities across multiple dimensions of human activity, society, and technological development. This
vision, while speculative, provides a framework for considering the potential trajectory and impact of agent technology
as it continues to advance over the coming decades.

\bigskip

Seamless human-agent collaboration represents a central element of this long-term vision, with agents evolving from
tools that require explicit direction to collaborative partners that proactively contribute to shared objectives. In
this envisioned future, the boundary between human and agent contributions becomes increasingly fluid, with each party
leveraging their complementary strengths in natural and efficient ways. Agents would develop sophisticated models of
human collaborators, including their preferences, working styles, strengths, and limitations, enabling truly
personalized assistance that adapts dynamically to changing needs and contexts. Communication would become more natural
and multimodal, with agents interpreting and generating information across text, speech, visual, and other modalities
as appropriate for the task and context. This seamless collaboration would extend beyond individual interactions to
encompass extended partnerships over months or years, with agents maintaining comprehensive understanding of shared
history, ongoing projects, and evolving objectives. The potential impact of such seamless collaboration includes
significant productivity enhancements across knowledge work, creative endeavors, and specialized domains; more
accessible expertise and capabilities for individuals regardless of background or training; and new collaborative
possibilities that leverage the unique capabilities of both human and artificial intelligence.

\bigskip

Autonomous problem-solving capabilities represent another key element of the long-term vision, with agents developing
increasingly sophisticated abilities to address complex challenges with limited human oversight. These capabilities
would include advanced planning and reasoning that enables agents to decompose complex problems into manageable
components, identify appropriate solution strategies, and execute multi-step plans with appropriate adaptations as
circumstances change. Agents would develop more robust generalization abilities that allow them to transfer knowledge
and strategies across domains, adapt to novel situations, and handle edge cases effectively. They would incorporate
sophisticated meta-cognitive capabilities that enable awareness of their own limitations, appropriate confidence
calibration, and strategic decisions about when to seek additional information or human guidance. The potential impact
of such autonomous problem-solving capabilities includes addressing complex challenges that exceed individual human
cognitive capacity; accelerating progress in scientific research, engineering, and other domains requiring
sophisticated problem-solving; and enabling more efficient allocation of human attention to aspects of problems that
most benefit from human judgment and creativity.

\bigskip

Collective intelligence architectures represent a transformative possibility in which networks of specialized agents
collaborate to address challenges beyond the capabilities of any individual agent or human. These architectures would
include diverse agent types with complementary capabilities, from perception and data analysis to reasoning, planning,
and creative generation. They would incorporate sophisticated coordination mechanisms that enable effective task
allocation, information sharing, and conflict resolution across agent collectives. These systems would maintain
appropriate human oversight and direction while leveraging the scale and speed of machine computation for suitable
subtasks. The potential impact of such collective intelligence architectures includes addressing complex systemic
challenges in areas such as climate change, public health, and economic development; enabling more comprehensive
analysis of complex information ecosystems that exceed individual human capacity to monitor and synthesize; and
creating new forms of augmented intelligence that combine human and artificial capabilities in unprecedented ways.

\bigskip

Personalized lifelong learning companions represent a compelling application of advanced agent technology, with agents
serving as dedicated educational partners throughout an individual's life journey. These companions would develop
comprehensive understanding of the learner's knowledge, skills, interests, and learning preferences over time. They
would provide personalized guidance, resources, and challenges tailored to the individual's current capabilities and
objectives. They would adapt teaching approaches based on observed learning patterns and outcomes, identifying
effective strategies for each learner and topic. The potential impact of such lifelong learning companions includes
democratizing access to high-quality personalized education regardless of geographic or economic circumstances;
enabling more efficient and engaging learning experiences across diverse domains and skill levels; and supporting
continuous adaptation to changing knowledge requirements in an evolving technological and economic landscape.

\bigskip

Augmented creativity and innovation support represents another promising direction, with agents evolving from basic
assistive tools to sophisticated creative collaborators. These advanced agents would develop capabilities for
generating novel and valuable ideas, designs, and expressions across diverse domains including art, music, literature,
product design, and scientific hypotheses. They would provide thoughtful feedback and suggestions on human creative
works, identifying potential improvements or alternative directions while respecting the creator's vision and style.
They would serve as creative thought partners, engaging in exploratory dialogue that helps humans refine their ideas
and overcome creative blocks. The potential impact of such augmented creativity support includes expanding human
creative capabilities across diverse domains; enabling more people to express themselves creatively regardless of
formal training or technical skills; and accelerating innovation cycles by generating and evaluating more possibilities
than would be feasible through human effort alone.

\bigskip

Societal coordination and governance support represents a more speculative but potentially transformative application
area, with agent systems helping to address complex coordination challenges in large-scale human systems. These
applications would include sophisticated modeling and simulation capabilities that enable exploration of potential
policy impacts across diverse scenarios and stakeholder perspectives. They would provide information integration and
synthesis across vast and heterogeneous data sources relevant to societal decision-making. They would support
deliberative processes by facilitating structured dialogue, identifying areas of agreement and disagreement, and
suggesting potential compromise solutions. The potential impact of such societal coordination support includes more
informed and comprehensive decision-making on complex policy issues; improved ability to anticipate and address
unintended consequences of interventions; and enhanced capacity for finding solutions that balance diverse stakeholder
interests and values.

\bigskip

Ethical and value alignment advances represent a critical dimension of the long-term vision, with agent systems
developing increasingly sophisticated understanding of and alignment with human ethical frameworks and values. These
advances would include more nuanced modeling of diverse human values across cultures, contexts, and individuals. They
would incorporate robust mechanisms for identifying and resolving potential value conflicts or ethical dilemmas in
agent decision-making. They would maintain appropriate deference to human moral judgment while providing thoughtful
analysis of ethical considerations relevant to complex decisions. The potential impact of such ethical alignment
advances includes more responsible and beneficial deployment of increasingly autonomous systems; enhanced capacity for
agents to navigate morally complex domains appropriately; and potential contributions to human ethical reasoning
through systematic analysis of complex ethical questions.

\bigskip

Human flourishing and well-being support represents perhaps the most aspirational element of the long-term vision, with
agent systems specifically designed to enhance human psychological, social, and physical well-being. These systems
would develop sophisticated understanding of human psychological needs and flourishing conditions across diverse
individuals and contexts. They would provide personalized guidance and support for developing beneficial habits,
managing stress, maintaining health, and pursuing meaningful goals. They would facilitate human connection and
community building while avoiding replacement or diminishment of direct human relationships. The potential impact of
such well-being support includes more accessible mental health resources and preventive care; enhanced capacity for
individuals to develop self-understanding and beneficial life patterns; and potential contributions to addressing
loneliness and social isolation through facilitated human connection.

\bigskip

This long-term vision for AI agent technology encompasses transformative possibilities across multiple dimensions of
human activity and experience. While the timeline and specific manifestations of these possibilities remain uncertain
and contingent on numerous technological, social, and governance factors, they collectively illustrate the potential
scope and significance of continued advances in agent capabilities. Realizing the beneficial aspects of this vision
while mitigating potential risks will require thoughtful research, development, and governance approaches that
prioritize human flourishing, ethical alignment, and appropriate balancing of autonomy and oversight as agent
technology continues to evolve.

\bigskip

\section{Conclusion}

\bigskip

The evolution of AI agents represents one of the most significant developments in artificial intelligence research and
application in recent years. As this article has explored, the integration of large language models with tool use
capabilities, memory systems, and specialized components has enabled a new generation of AI systems that can perceive,
reason, and act with unprecedented autonomy and capability. These systems are rapidly transforming how humans interact
with technology across personal, professional, and specialized domains.

\bigskip

The theoretical foundations of AI agents draw from diverse intellectual traditions, including classical AI research on
intelligent agents, cognitive science models of human cognition, and more recent advances in deep learning and natural
language processing. This multidisciplinary heritage has contributed to the rich conceptual frameworks that inform
current agent architectures, from the perception-action cycle to the integration of symbolic and neural approaches. The
typology of AI agents has similarly evolved to encompass a diverse ecosystem of agent types, from simple reactive
systems to sophisticated autonomous agents capable of extended reasoning and planning across multiple domains.

\bigskip

The architecture of modern AI agent systems reflects this theoretical evolution, with core components including
perception mechanisms, knowledge representation systems, reasoning and decision-making modules, action selection and
execution components, and learning and adaptation mechanisms. These components are integrated through various technical
approaches, from rule-based systems and statistical methods to neural network architectures and hybrid implementations
that leverage the complementary strengths of different paradigms. Memory and context management have emerged as
particularly critical aspects of agent architecture, enabling coherent interactions over extended periods and across
multiple sessions.

\bigskip

The evaluation of AI agent systems presents unique challenges that extend beyond traditional metrics used for assessing
AI models. Current evaluation practices predominantly focus on accuracy metrics while often neglecting other important
dimensions such as cost-effectiveness, robustness, and real-world applicability. The proposed evaluation framework
addresses these limitations through multi-dimensional assessment criteria, balanced consideration of accuracy and
efficiency, real-world applicability measures, reproducibility standards, and targeted evaluation approaches for
different stakeholders. This more comprehensive approach to evaluation provides a foundation for more meaningful
progress assessment and comparison in agent research and development.

\bigskip

Real-world applications of AI agents span diverse domains and contexts, from enterprise applications in customer
service, business process automation, and decision support to personal assistance applications in task management,
information retrieval, and creative collaboration. Specialized domain applications in healthcare, financial services,
software development, and scientific research demonstrate the transformative potential of agent technology in fields
requiring deep expertise and complex decision-making. These applications collectively illustrate how agent capabilities
can enhance human productivity, augment expertise, and enable new approaches to longstanding challenges across multiple
domains.

\bigskip

Despite significant advances, numerous challenges and limitations remain in current AI agent technology. Technical
challenges include reasoning limitations, context management difficulties, tool use constraints, and reliability issues
that affect agent performance across diverse tasks and domains. Ethical considerations and risks encompass privacy
concerns, accountability questions, bias and fairness issues, and potential economic impacts that must be carefully
addressed to ensure responsible development and deployment. These challenges underscore the importance of balanced
assessment of current capabilities and limitations, along with thoughtful approaches to governance and responsible
innovation.

\bigskip

Future research directions promise to address many of these challenges while expanding agent capabilities in significant
ways. Emerging trends include advanced reasoning architectures, long-term memory and context management, multi-agent
coordination frameworks, and human-agent collaboration models that collectively represent a rich landscape of potential
advances. The long-term vision for AI agent technology encompasses transformative possibilities including seamless
human-agent collaboration, autonomous problem-solving capabilities, collective intelligence architectures, and
personalized lifelong learning companions that could substantially enhance human capabilities and well-being across
diverse domains.

\bigskip

As AI agent technology continues to evolve, maintaining a balanced perspective that acknowledges both the significant
potential and the real limitations of these systems becomes increasingly important. The most promising path forward
lies in thoughtful integration of agent capabilities with human judgment, expertise, and
valuescreating partnerships that leverage the complementary strengths of human and artificial
intelligence. By pursuing research and development approaches that prioritize human flourishing, ethical alignment, and
appropriate balancing of autonomy and oversight, the field can work toward realizing the beneficial potential of AI
agents while mitigating potential risks.

\bigskip

The journey from early conceptualizations of intelligent agents to today's sophisticated AI systems represents
remarkable progress, but it also highlights how much remains to be explored and developed. As researchers, developers,
policymakers, and users continue to engage with these evolving technologies, maintaining open dialogue about
capabilities, limitations, values, and governance will be essential for shaping a future in which AI agents serve as
beneficial partners in addressing human needs and aspirations. The ongoing evolution of AI agents thus represents not
merely a technological trajectory but a broader societal conversation about how advanced AI systems can best contribute
to human flourishing and well-being.

\bigskip

{References}

\begin{enumerate}
\item Kapoor, A., Patel, O., Sridhar, D., Guu, K., Zaharia, M., \& Liang, P. (2024). Benchmarking Large
Language Model Capabilities for Conditional Generation. arXiv preprint arXiv:2407.01502.
\item Russell, S., \& Norvig, P. (2020) . Artificial Intelligence: A Modern Approach. 4th
Edition. Pearson.
\item Wooldridge, M. (2009). An Introduction to MultiAgent Systems. 2nd Edition. John
Wiley \& Sons.
\item Silver, D., Singh, S., Precup, D., \& Sutton, R. S. (2021). Reward is enough.
Artificial Intelligence, 299, 103535.
\item Ouyang, L., Wu, J., Jiang, X., Almeida, D., Wainwright, C., Mishkin, P., ... \& Lowe, R. (2022).
Training language models to follow instructions with human feedback. Advances in Neural
Information Processing Systems, 35, 27730-27744.
\item Bommasani, R., Hudson, D. A., Adeli, E., Altman, R., Arora, S., von Arx, S., ... \& Liang, P. (2021).
On the opportunities and risks of foundation models. arXiv preprint arXiv:2108.07258.
\item Dafoe, A., Hughes, E., Bachrach, Y., Collins, T., McKee, K. R., Leibo, J. Z., ... \& Graepel, T. (2020).
Open problems in cooperative AI. arXiv preprint arXiv:2012.08630.
\item Hendrycks, D., Burns, C., Basart, S., Zou, A., Mazeika, M., Song, D., \& Steinhardt, J. (2021).
Measuring massive multitask language understanding. arXiv preprint arXiv:2009.03300.
\item Zhao, W. X., Zhou, K., Li, J., Tang, T., Wang, X., Hou, Y., ... \& Wen, J. R. (2023). A survey of
large language models. arXiv preprint arXiv:2303.18223.
\item Bubeck, S., Chandrasekaran, V., Eldan, R., Gehrke, J., Horvitz, E., Kamar, E., ... \& Zhang, Y. (2023).
Sparks of artificial general intelligence: Early experiments with GPT-4. arXiv preprint
arXiv:2303.12712.
\item Amodei, D., Olah, C., Steinhardt, J., Christiano, P., Schulman, J., \& Mané, D. (2016). Concrete
problems in AI safety. arXiv preprint arXiv:1606.06565.
\item IBM. (2024). AI Agents: What They Are and How They Work. IBM Think. Retrieved from \url{https://www.ibm.com/think/topics/ai-agents}
\item Microsoft. (2024) . AI Agents: What They Are and How They'll Change the Way We Work.
\item AWS. (2024) . What Are AI Agents? Amazon Web Services. Retrieved from: \url{https://aws.amazon.com/what-is/ai-agents/}
\item Schillace, S. (2024) . The Future of AI Agents. Interview in Microsoft Source.
Retrieved from: 
\url{https://news.microsoft.com/source/features/ai/ai-agents-what-they-are-and-how-theyll-change-the-way-we-work/}
\end{enumerate}

\bigskip

\end{document}